\documentclass[10pt,twocolumn,letterpaper]{article}

\usepackage[pagenumbers]{cvpr} 

\usepackage[table]{xcolor}
\usepackage{multirow}
\usepackage{float}

\definecolor{cvprblue}{rgb}{0.21,0.49,0.74}
\usepackage[pagebackref,breaklinks,colorlinks,allcolors=cvprblue]{hyperref}

\title{PackCache: A Training-Free Acceleration Method for Unified Autoregressive Video Generation via Compact KV-Cache}

\author{
Kunyang Li, Mubarak Shah, Yuzhang Shang \\
Institute of Artificial Intelligence, University of Central Florida \\
}

\begin{document}
\maketitle
\nocite{*}

\begin{abstract}
A unified autoregressive model is a Transformer-based framework that addresses diverse multimodal tasks (e.g., text, image, video) as a single sequence modeling problem under a shared token space. Such models rely on the KV-cache mechanism to reduce attention computation from $\mathcal{O}(T^{2})$ to $\mathcal{O}(T)$; however, KV-cache size grows linearly with the number of generated tokens, it rapidly becomes the dominant bottleneck limiting inference efficiency and generative length. Unified autoregressive video generation inherits this limitation. Our analysis reveals that KV-cache tokens exhibit distinct spatiotemporal properties: (i) text and conditioning-image tokens act as persistent semantic anchors that consistently receive high attention, and (ii) attention of previous frames naturally decays with temporal distance. Leveraging these observations, we introduce PackCache, a training-free KV-cache management method which dynamically compacts the KV cache through three coordinated mechanisms: condition anchoring that preserves semantic references, cross-frame decay
modeling that allocates cache budget according to temporal distance, and spatially preserving position embedding that
maintains coherent 3D structure under cache removal. 
In terms of efficiency, PackCache accelerates end-to-end generation by 1.7–2.2× on 48-frame long sequences showcasing its strong potential for enabling longer-sequence video generation. Notably, the final four frames— the portion most impacted by the progressively expanding KV-cache and thus the most expensive segment of the clip—PackCache delivers a 2.6× and 3.7× acceleration on A40 and H200, for 48-frame videos.




\end{abstract}

\begin{figure}[t]
  \centering
  \includegraphics[width=\linewidth]{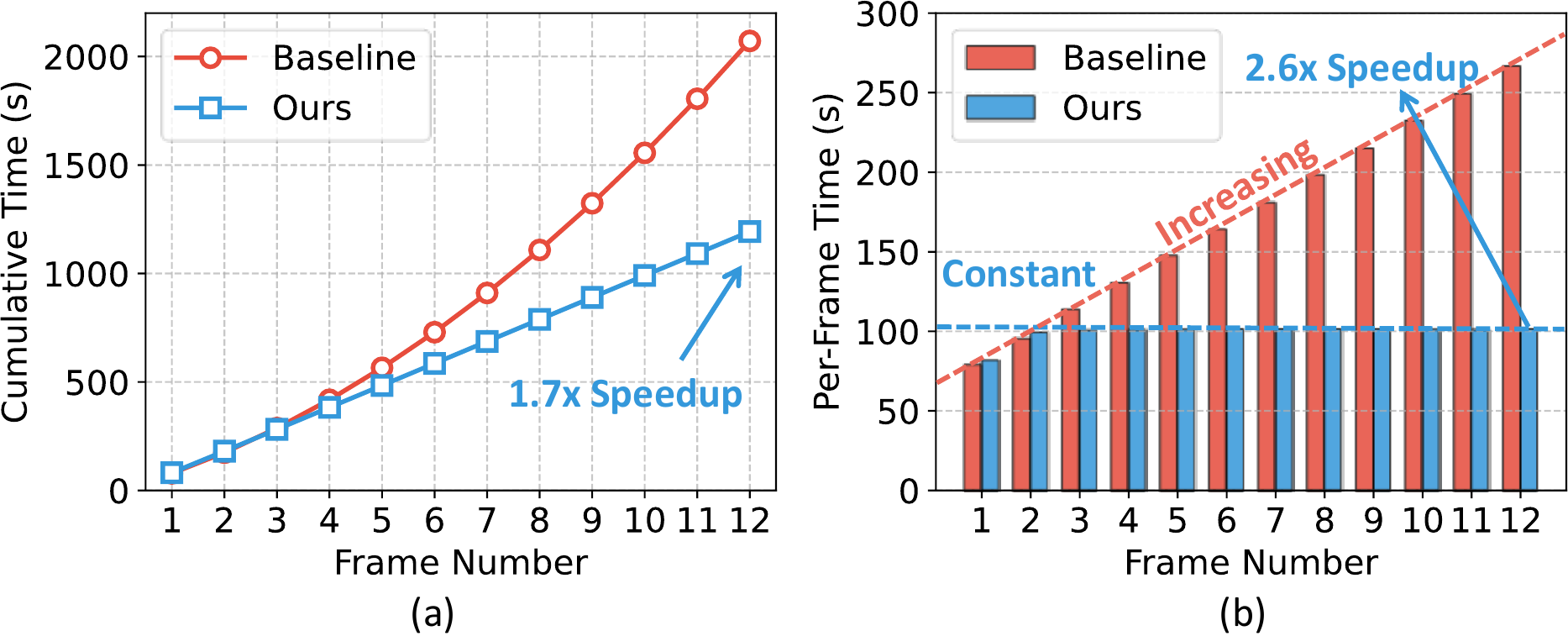}
  \caption{24-Frame generation overhead of unified autoregressive video models on an NVIDIA A40 GPU. \textbf{(a)} Cumulative generation time for a 24-frame video ($384\times672$) using the 3B-parameter Lumos-1. \textbf{(b)} Per-frame generation time. The cumulative curve exhibits rapidly increasing trend 
  as the sequence length increases, and the incremental cost of later frames rises to over $160$ seconds per frame, indicating severe KV-cache–driven scaling bottlenecks. Our method accelerates end-to-end generation by $1.7\times$ compared to the baseline, and achieves a $2.6\times$ speedup on the final four frames when generating 48-frame videos on the A40 GPU.}
  \label{fig:overhead}
\end{figure}

\section{Introduction}
\label{sec:intro}

Advances in large language models (LLMs) have demonstrated that a unified autoregressive paradigm, formulated as next-token prediction, can generalize across diverse linguistic tasks \cite{raffel2020exploring, brown2020language, chowdhery2023palm, achiam2023gpt, touvron2023llama}. Inspired by the this paradigm, recent visual generation studies have increasingly integrated autoregressive modeling into diffusion-based frameworks\cite{yin2025slow,jin2024pyramidal,teng2025magi}, leveraging its strength in long-range temporal dependency modeling. Meanwhile, treating the entire spatiotemporal token sequence as a single context and generating videos end-to-end with one autoregressive Transformer\cite{vaswani2017attention} has also proven effective\cite{yu2024languagemodelbeatsdiffusion,jin2025pyramidalflowmatchingefficient,yu2025videomarautoregressivevideogeneratio,wang2025loonggeneratingminutelevellong,villegas2022phenakivariablelengthvideo,kondratyuk2024videopoetlargelanguagemodel,yuan2025lumos}, further validating the potential of unified autoregressive approaches for video generation.

In video generation models built upon LLM architectures, key–value pairs from previous generated frames are cached and reused by queries of the current frame to improve inference efficiency. By reusing cached keys and values, the attention computation per decoding step is reduced from quadratic to linear complexity with respect to sequence length $T$—i.e., from $\mathcal{O}(T^2)$ to $\mathcal{O}(T)$ per step—thereby avoiding redundant recomputation. However, as generation progresses, the KV cache itself grows linearly with the number of frames, making attention computation a new bottleneck due to the rapidly increasing memory footprint and latency cost~\cite{li2024survey, ge2023model,tang2024quest, xiao2024efficient}.

\begin{figure}[t]
  \centering
  \includegraphics[width=\linewidth]{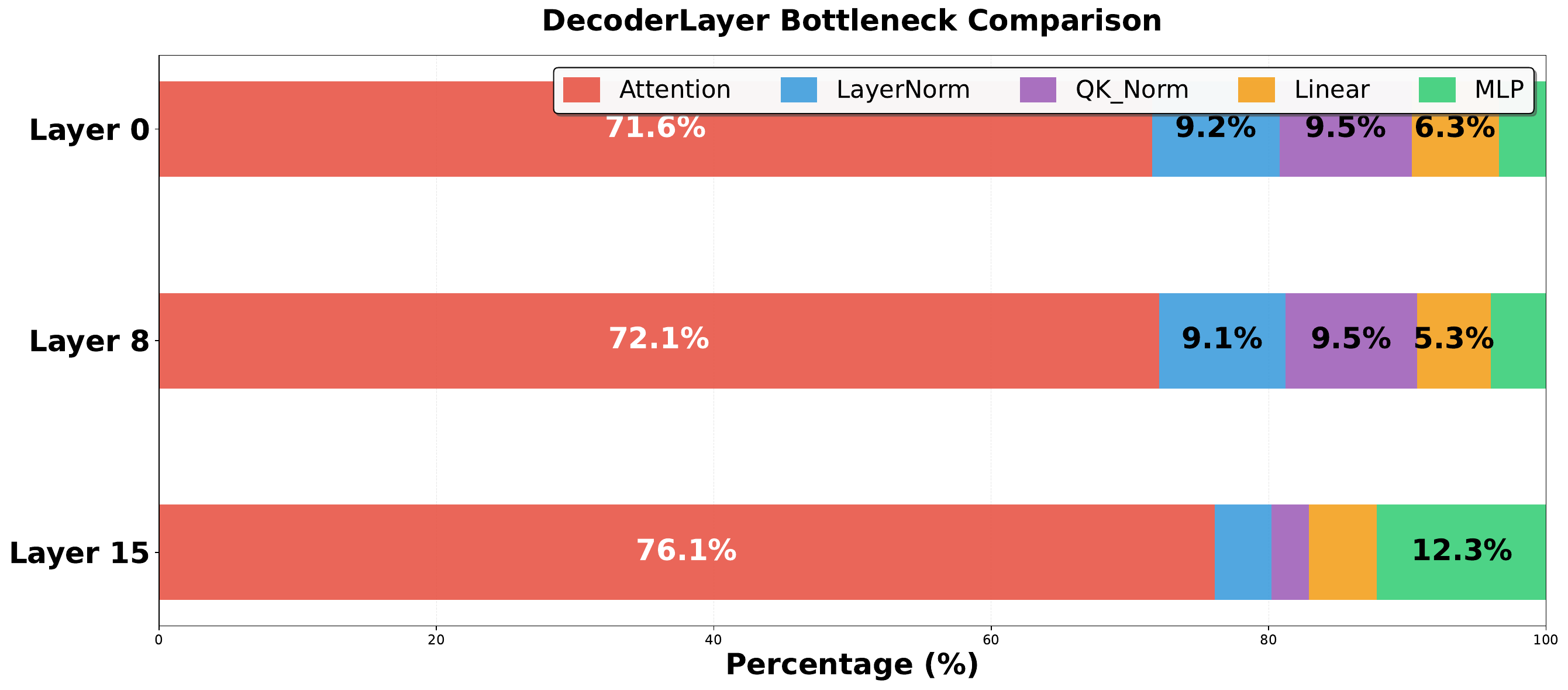}
      \vspace{-0.2in}
  \caption{DecoderLayer Bottleneck Comparison Relative time distribution of module-level operations in three decoder layers. Attention dominates computation (71-76\%), with LayerNorm and QK\_Norm contributing ~9\% each. Values represent percentage of total layer execution time, measured with GPU synchronization over 612 forward passes.}
      \vspace{-0.2in}
  \label{fig:bottleneck}
\end{figure}

Inference costs for long-context Transformers (e.g., 100K–1M tokens) are considerably higher than for short-context variants (e.g., 4K tokens), with most of the overhead arising from the expanding KV cache \cite{fu2024challenges}. Hardware scaling (RTX~4090 $\rightarrow$ A100 $\rightarrow$ H100) only reduces latency linearly, which cannot close the gap between 50K and 4K contexts. This issue is even more pronounced in autoregressive video generation, as generating longer sequences steadily grows the KV-cache. Discretizing a 48-frame $384\times672$ video with the Cosmos Tokenizer\cite{nvidia2025cosmosworldfoundationmodel}, which encodes the video into its latent space, produces 13 latent frames,
resulting in roughly $\sim$53K tokens—already exceeding the 50K long-context threshold. Empirically, using 3B-parameter Lumos-1 on an A40, generating a 24-frame video already takes over 12 minutes, and longer sequences cause both latency and memory to escalate rapidly (See Fig. \ref{fig:overhead}). Module-level profiling (See Fig.~\ref{fig:bottleneck}) further reveals that Attention is the dominant bottleneck, accounting for 71–76\% of per-layer computation, while all other components together contribute only 24–29\%.

To address the above challenges, many studies on large langugage models (LLMs) have explored KV cache management to accelerate inference by reducing redundant computations and improving memory utilization\cite{li2024survey}. For instance, MQA\cite{shazeer2019fasttransformerdecodingwritehead} and GQA\cite{ainslie2023gqatraininggeneralizedmultiquery} enhance KV-cache efficiency through structural sharing—MQA lets all attention heads share the same key–value pairs, while GQA further divides query heads into groups, each maintaining its own shared K/V set. However, these architectural modifications usually require retraining or fine-tuning the whole models, making the realization costly. As opposed to this, several approaches \cite{li2024snapkv, chen2024sepllm,liu2025clusterkv,cai2024pyramidkv,wang2024model,sheng2023flexgen,shao2023omniquant} take on token-level selection, essentially managing the KV cache during inference in a training-free manner. For example, FastGen\cite{ge2023model} performs pattern-aware static KV selection during the prefill stage; H2O\cite{zhang2023h2o} dynamically selects high-impact tokens based on attention scores; and Quest\cite{tang2024quest} achieves efficient non-permanent dynamic selection via block-level indexing.

\begin{figure*}
  \centering
  
  \begin{subfigure}{0.36\linewidth}
  \includegraphics[width=1.0\linewidth]{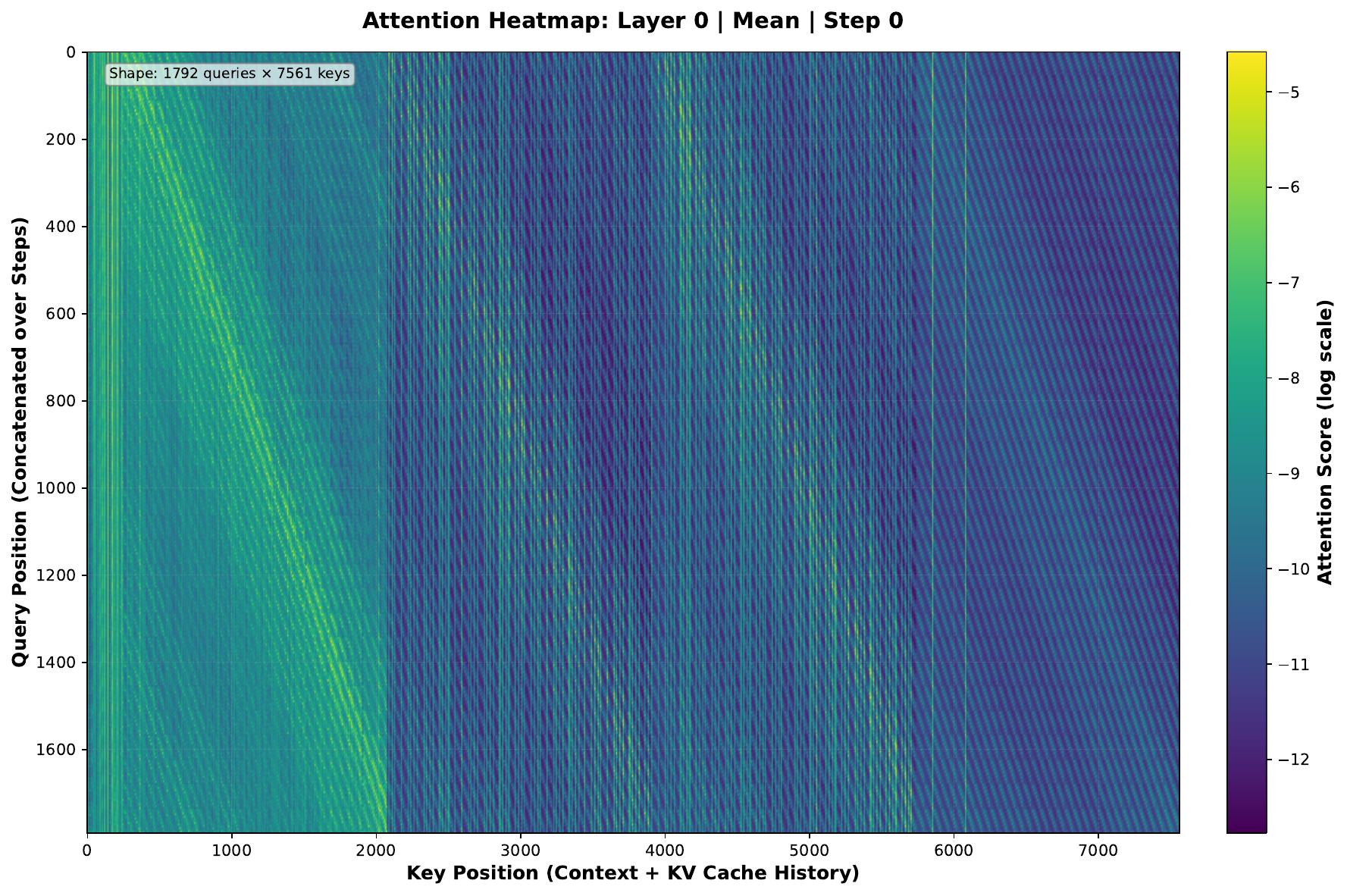}
    \caption{Layer~0 heatmap at Step~0, showing strong attention to conditioning inputs.}
    \label{fig:short-a}
  \end{subfigure}
  \hfill
  \begin{subfigure}{0.3\linewidth}
  \includegraphics[width=0.9\linewidth]{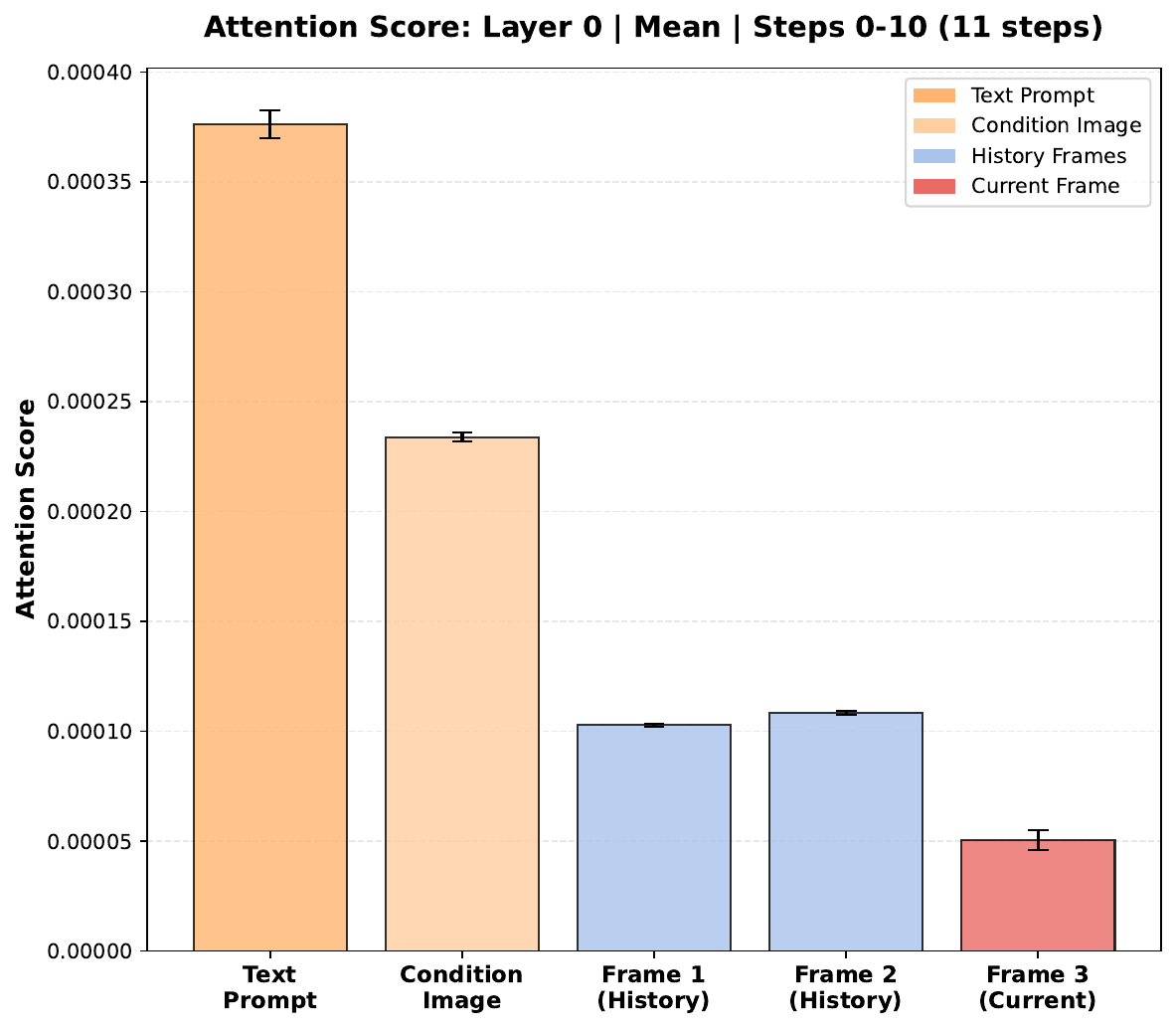}
    \caption{Layer~0 attention (Steps~0--10), dominated by prompt and conditioning image. }
    \label{fig:short-b}
  \end{subfigure}
  \begin{subfigure}{0.3\linewidth}
  \includegraphics[width=0.9\linewidth]{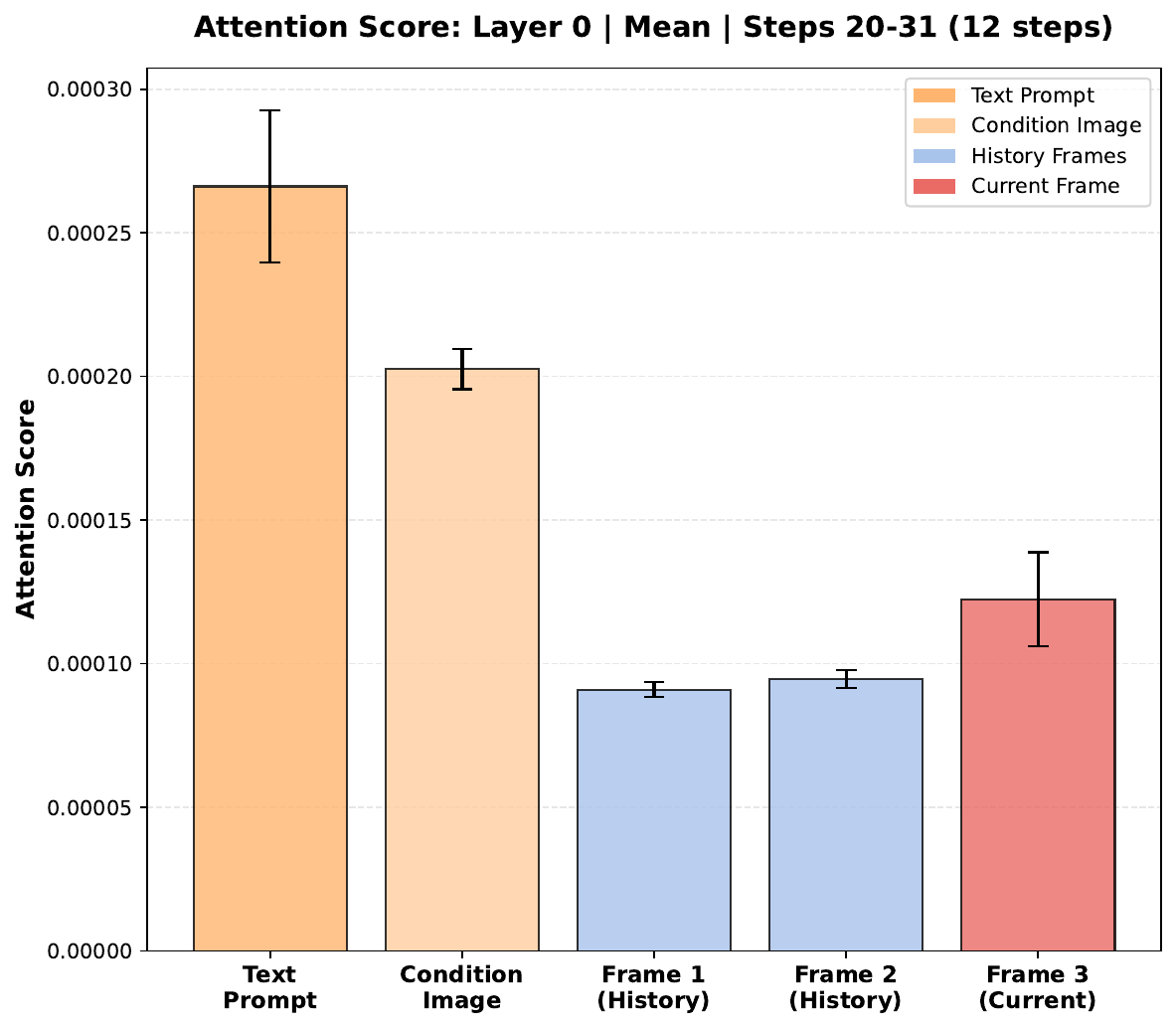}
    \caption{Layer~0 attention (Steps~20--31), increasing focus on recent frames. }
    \label{fig:short-c}
  \end{subfigure}

    \begin{subfigure}{0.36\linewidth}
  \includegraphics[width=1.0\linewidth]{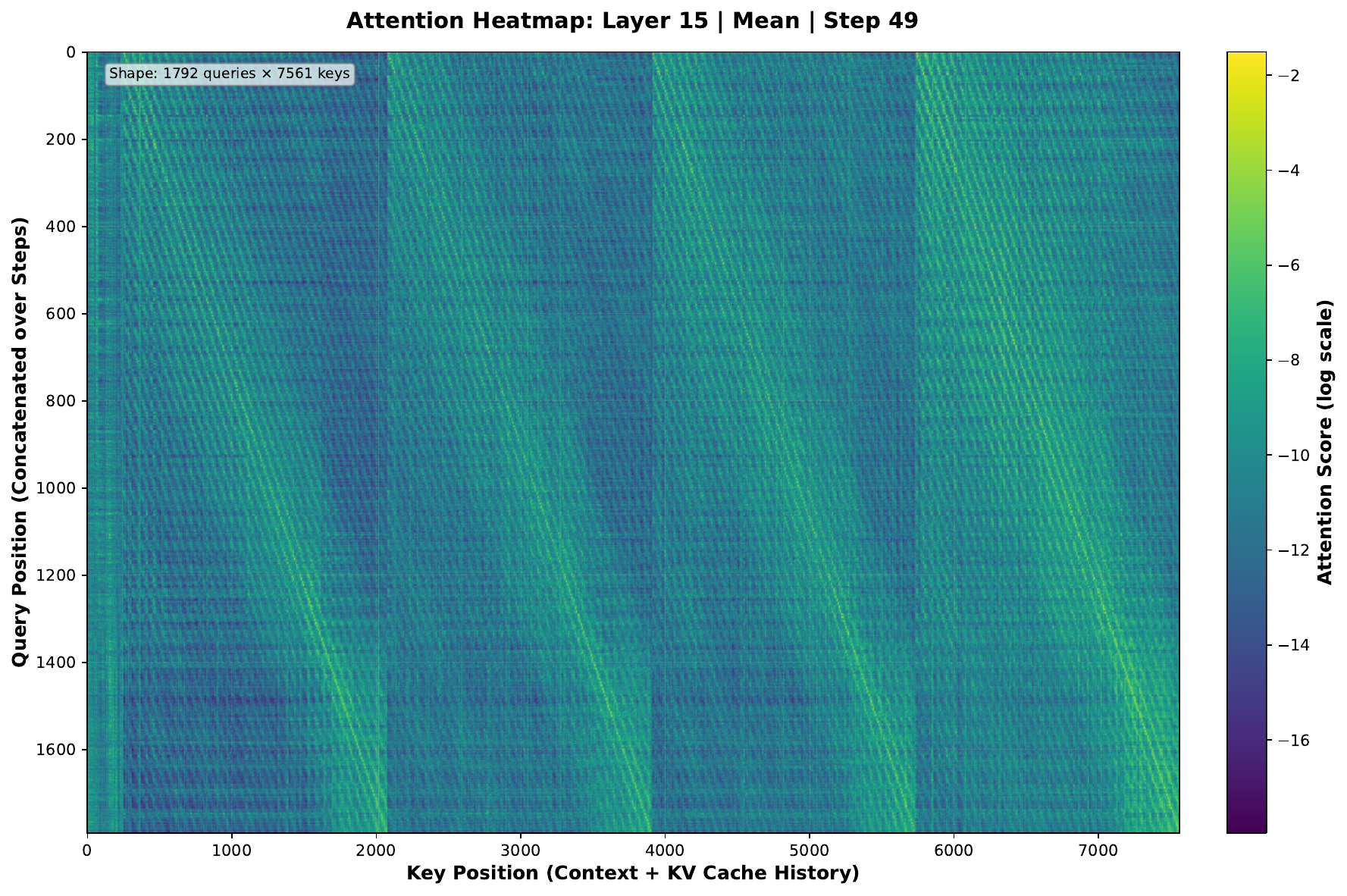}
    \caption{Layer~15 heatmap at Step~49, exhibiting clearer temporal decay. }
    \label{fig:short-d}
  \end{subfigure}
  \hfill
  \begin{subfigure}{0.3\linewidth}
  \includegraphics[width=0.9\linewidth]{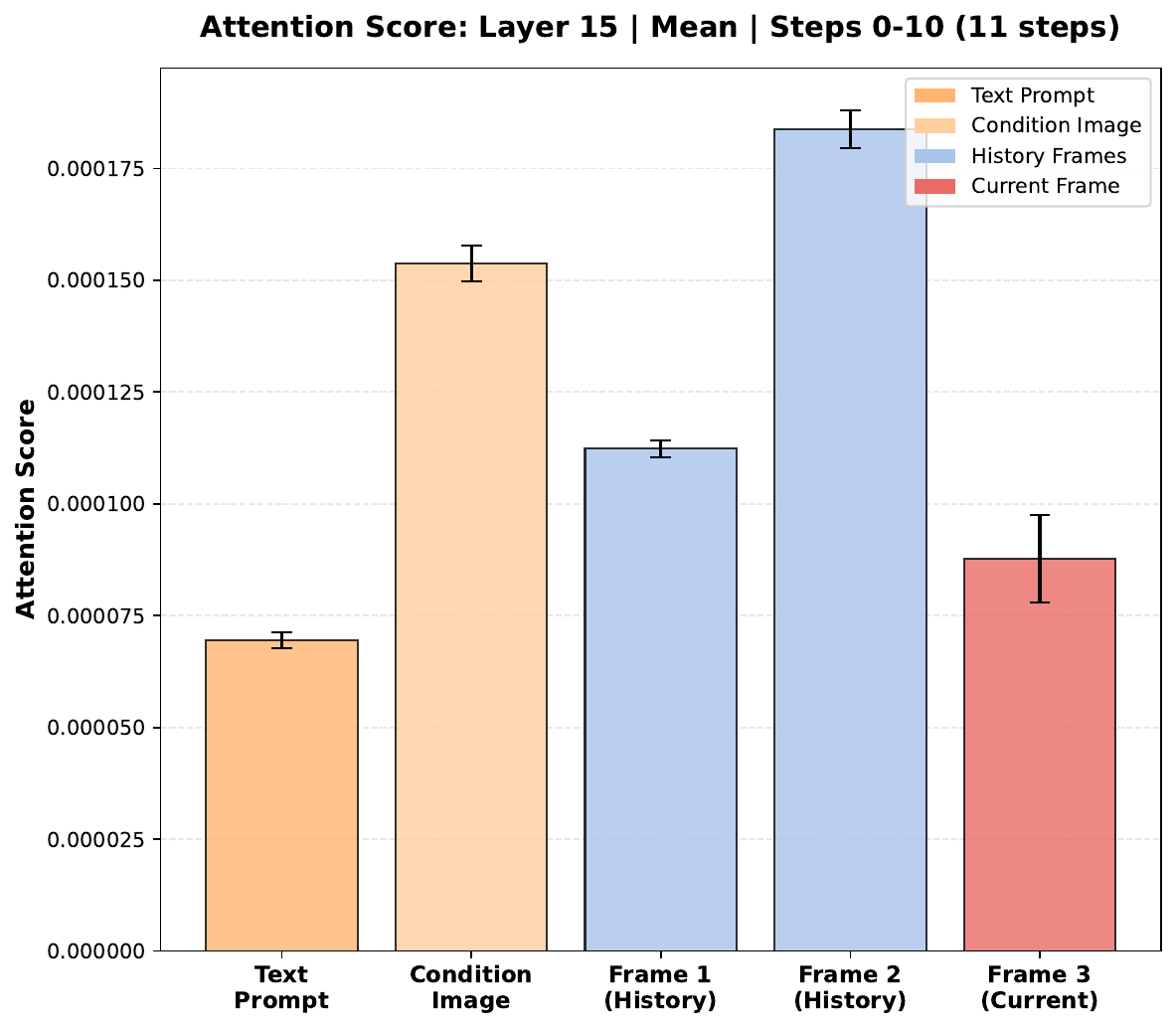}
    \caption{Layer~15 attention ($<$Steps~10), showing early-stage temporal structure.  }
    \label{fig:short-e}
  \end{subfigure}
  \begin{subfigure}{0.3\linewidth}
  \includegraphics[width=0.9\linewidth]{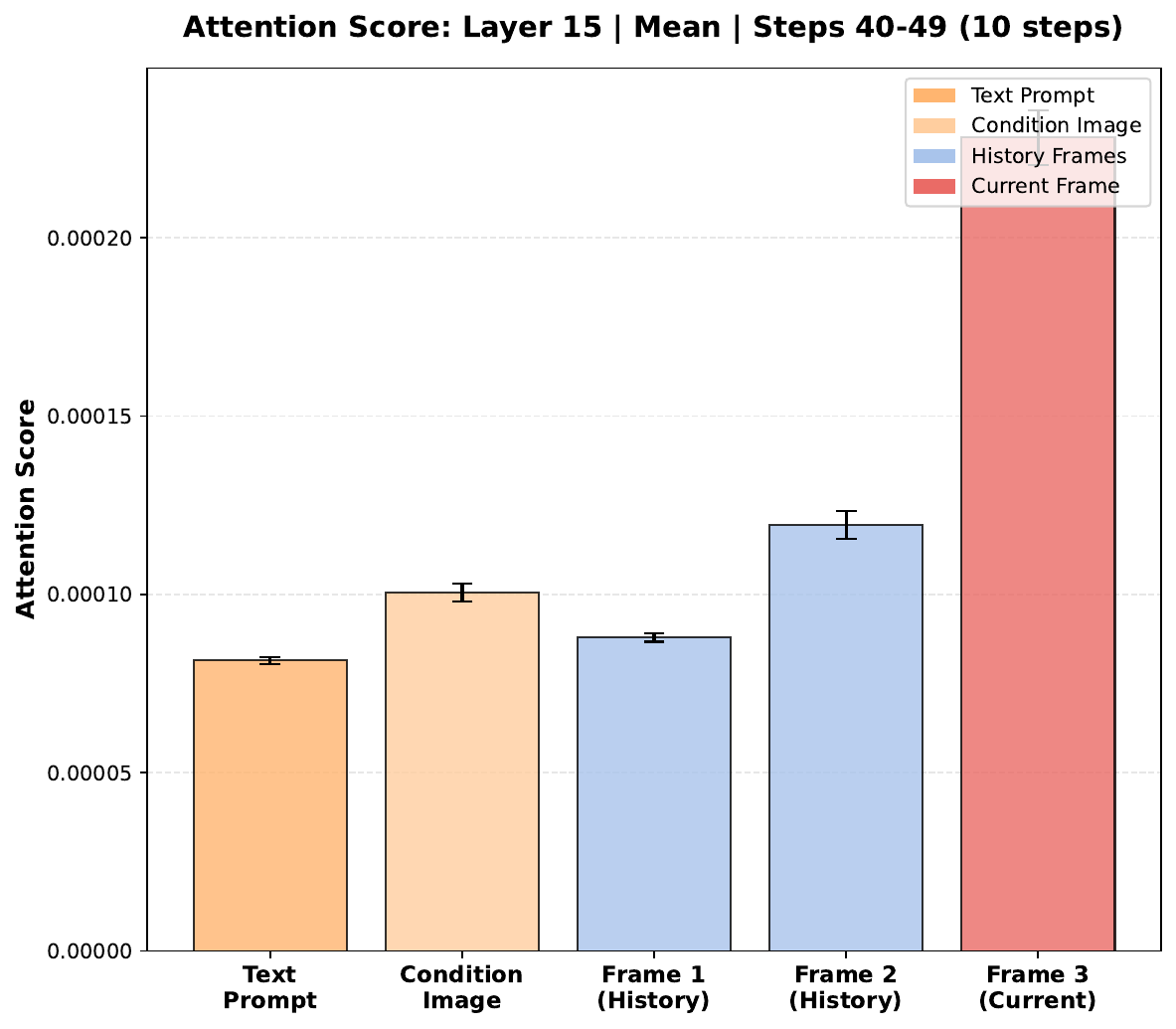}
    \caption{Layer~15 attention (40--49), emphasizing the current frame and persistent conditioning.}
    \label{fig:short-f}
  \end{subfigure}
  
  \caption{Attention heatmap and attention-distribution visualizations of the 1B Lumos-1 model during the generation of latent frame~3. In the heatmaps (a,d), the horizontal axis enumerates KV-cache key tokens: \([0,249)\) corresponds to text-prompt tokens, \([249,2077)\) to the conditioning image, \([2077,3905)\) to the earliest cached frame, \([3905,5733)\) to the second-most recent frame, and \([5733,7561)\) to the masked current-frame tokens. The vertical axis represents the KV-cache query positions of the current frame. Each heatmap shows a single timestep from a selected layer, averaged over all attention heads.Autoregressive generation proceeds by iteratively predicting masked tokens over multiple timesteps. The attention-distribution plots (b,c,e,f) aggregate the mean attention scores (averaged across heads) from current-frame queries to each semantic region in the cache. These visualizations reveal three consistent patterns: (1) a rightward shift in attention toward more recent frames, (2) a monotonic temporal decay across history frames, and (3) strong, persistent attention to the text prompt and conditioning image, which act as stable semantic anchors throughout autoregressive video synthesis.
}

  \vspace{-0.15in}
  \label{fig:heatmap}
\end{figure*}

However, in autoregressive video modeling, KV cache management for visual tokens remains largely unexplored. Language tokens are highly abstract, context-aware, and encoded with one-dimensional positional embeddings, whereas vision tokens exhibit a certain degree of sparsity \cite{zhang2025frame} and rely on three-dimensional positional encoding \cite{yuan2025lumos}. Through analyzing attention heatmaps~\ref{fig:heatmap} across layers and generation steps, we reveal two key properties of KV-cache behavior in autoregressive video generation models: (1) tokens in the current frame assign higher attention weights to temporally closer frames than to distant ones, and (2) both the text prompt and the first conditioning image consistently receive strong attention, functioning as stable semantic anchors throughout generation. 

Motivated primarily by above two properties, we introduce a training-free KV-cache token compaction method tailored for autoregressive video generation. Our approach first preserves the semantic anchors (text prompt and conditioning image) in the cache, then applies a temporal distance–aware compaction strategy, retaining a larger fraction of tokens from closer frames and fewer from distant ones. This keeps the KV cache within a fixed budget while preserving substantially longer historical context than sliding-window strategy, which only retain the most recent frame in the cache. Furthermore, reflecting the positional-embedding differences between language and vision tokens, our compaction strategy keeps the 1D temporal positional embeddings continuous to preserve temporal coherence, while leaving the 3D spatialtemporal positional embeddings unchanged to maintain spatial consistency. As a result, our method reduces inference cost with minimal overhead while achieving significant speedups with minimal degradation in video quality.
Our key contributions are as follows:
\begin{itemize}
    \item We conduct a systematic analysis of attention behaviors in unified video modeling with KV cache, revealing that the text prompt and conditioning image act as semantic anchors with consistently high attention, while attention strength decays with increasing temporal distance.
    \item We propose PackCache, a training-free method that dynamically compacts the KV cache with minimal overhead. By maintaining a fixed cache size while preserving longer context, PackCache effectively reduces temporal inconsistencies 
    during long video generation. Combined with the Spatially Preserving Position Embedding, it further enables coherent long-range video synthesis.

    \item PackCache delivers substantial efficiency gains: it accelerates 24-frame generation by 1.3--1.5$\times$ (1.6--1.7$\times$ on the final four frames), and achieves 1.7--2.2$\times$ speed-up on 48-frame videos, with up to 2.6--3.7$\times$ improvement for the most expensive final frames.

\end{itemize}

\section{Related Work}
\label{sec:relate}

\subsection{Unified Autoregressive Video Modeling}
Advances in large-scale generative modeling have given rise to unified models—frameworks capable of handling multiple modalities or tasks within a single network architecture. Instead of designing modality-specific branches, unified models aim to represent text, image, video, and even audio through a shared tokenization and autoregressive or diffusion-based modeling paradigm. Representative examples include Unified-IO-2 \cite{lu2024unified}, SEED-X \cite{ge2024seed}, Kosmos-2 \cite{peng2023kosmos}, Florence-2\cite{xiao2024florence}, Vargpt \cite{zhuang2025vargpt} and Chameleon \cite{team2024chameleon}, all of which seek to unify understanding and generation under a single multimodal Transformer backbone. 
From this unified-model perspective, Lumos-1\cite{yuan2025lumos} is the first open-sourced unified video generation model. Rather than designing modality-specific architectures or task-dependent objectives, Lumos-1\cite{yuan2025lumos} retains the original LLM \cite{touvron2023llama} architecture with minimal yet principled modifications: MM-RoPE, which injects balanced spatiotemporal correlations via distributed and scaled 3D rotary position embeddings; and AR-DF (Autoregressive Discrete Diffusion Forcing), which enforces temporally causal yet spatially bidirectional dependencies. These innovations enable Lumos-1\cite{yuan2025lumos} to perform text-to-image, image-to-video, and text-to-video generation within a single autoregressive framework, demonstrating the feasibility of scaling LLMs into general-purpose multimodal generators.
Building upon the unified modeling paradigm, our work is targeted at acceleration of unified video generation models with the compact KV-Cache. 

\begin{figure}[t]
  \centering
  \includegraphics[width=\linewidth]{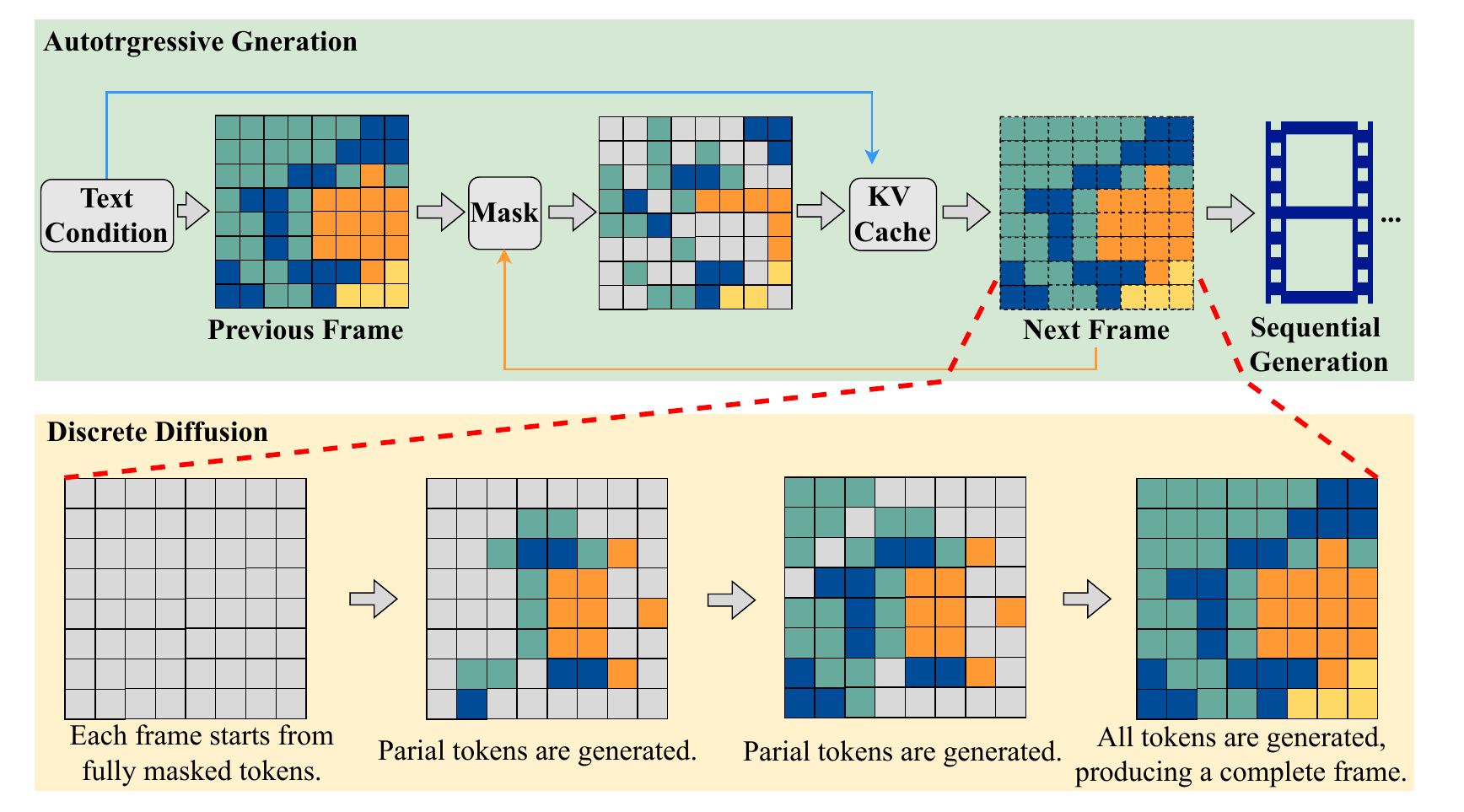}
      \vspace{-0.2in}
  \caption{The AR-DF pipeline in Lumos-1\cite{yuan2025lumos}. In the top row, video frames are generated autoregressively, where each frame is conditioned on text prompts and partial visual tokens from previous frames stored in the KV cache. \textcolor[HTML]{3399FF}{The text prompt and condition image are encoded and stored in the KV cache as global context. \textcolor[HTML]{FF9933}{The partially visible tokens are written into the KV cache as contextual information for subsequent frames.}}
 In the bottom row, each frame is synthesized via discrete diffusion, progressively replacing masked tokens with valid visual tokens until a complete frame is obtained.}
      \vspace{-0.2in}
  \label{fig:ardf}
\end{figure}

\subsection{KV Cache Management}
KV Cache Management aims to balance computational efficiency and memory utilization during large language models (LLMs) \cite{touvron2023llama,radford2019language,dai2024deepseekmoe,zeng2022glm} inference by optimizing how key-value pairs are stored, updated, and reused across decoding steps\cite{li2024survey}. Specifically, token-level optimization \cite{ge2023model,li2024snapkv,zhang2023h2o, chen2024sepllm,liu2025clusterkv,cai2024pyramidkv,wang2024model,sheng2023flexgen,shao2023omniquant} focuses on fine-grained manipulation of individual tokens’ KV pairs, including selection, allocation, merging, quantization, and low-rank decomposition, thereby reducing memory consumption and inference latency without requiring any modification to the model architecture. Model-level optimization \cite{shazeer2019fast,ainslie2023gqa,brandon2024reducing,hua2022transformer,liu2024deepseek}redesigns internal architectures—such as grouped or shared attention and memory-efficient transformer variants—to structurally reduce KV redundancy and improve reuse efficiency. System-level optimization \cite{ye2024chunkattention,gao2025apt,xiong2024layerkv,juravsky2024hydragen,he2024fastdecode,pan2024instinfer} enhances runtime performance through hardware-aware memory management and scheduling, enabling scalable, high-throughput inference across heterogeneous devices. Unified video modeling, grounded in large language model (LLM) architectures, also employs KV cache to reduce the attention computation cost during video generation. While extensive research has addressed KV cache management in LLMs, the management of vision tokens requires further exploration. Unlike language tokens—highly abstract, context-aware, and encoded with one-dimensional positional embeddings—vision tokens have lower information density and adopt three-dimensional positional encoding. Thus, language tokens favor semantic-importance-based selection, whereas vision tokens call for spatiotemporal-structure-aware strategies. Our work focuses on a training-free approach that controls KV cache size through token selection, aiming to accelerate video generation while maintaining quality with minimal computational overhead.

\section{Method}
\label{sec:method}
In this section, we present PackCache, a training-free KV-cache management method designed to accelerate unified autoregressive (AR) video generation. We begin in Sec.\ref{subsec:pre} by reviewing the attention mechanism and position embedding scheme in autoregressive video models, locating the computational bottleneck posed by linearly growing KV caches. 
In Sec.\ref{subsec:analysis}, we systematically analyze attention attribution patterns across temporal frames, revealing two key properties: (i) persistent high attention to text prompts and conditioning images, and (ii) gradual temporal decay in cross-frame attention. 
We also examine the limitations of naive sliding-window strategies, which sacrifice temporal context for bounded memory in Sec.\ref{subsec:limitation}. 
Building on these insights, we introduce our PackCache framework in Sec.\ref{subsec:packcache}, which dynamically compacts the KV cache through three coordinated mechanisms: condition anchoring that preserves semantic references, cross-frame decay modeling that allocates cache budget according to temporal distance, and spatial-preserving position embedding that maintains coherent 3D structure under cache removal. 
This combination of components is capable of generating stable long sequences under strict memory constraints while preserving visual quality.

\subsection{Preliminary: Unified AR Video Generation}
\label{subsec:pre}

Autoregressive video generators following the Autoregressive Discrete Diffusion Forcing (AR-DF) framework~\cite{yuan2025lumos} expose each frame to only partial observations of earlier frames during training. To maintain this train--test consistency at inference, a frame-level Bernoulli mask determines which subset of previously generated tokens is written into the KV cache and thus remains visible to future frames. Given cached keys $ K_{\text{cache}} \in \mathbb{R}^{T\times N\times d_k}$, values $\qquad 
V_{\text{cache}} \in \mathbb{R}^{T\times N\times d_v}$ and a binary cache mask $M_{\text{cache}} \in \{0,1\}^{T\times N}$, the effective cache becomes
\begin{equation}
K = K_{\text{cache}} \odot M_{\text{cache}}, \qquad
V = V_{\text{cache}} \odot M_{\text{cache}},
\label{eq:cache_mask_single}
\end{equation}
where masked entries are ignored in attention but still occupy memory.  
During inference, AR-DF initializes each frame with masked tokens, applies a temporal causal mask, predicts the next latent frame, and caches only its unmasked subset. This ensures consistent partial observability across timesteps.

To encode spatiotemporal structure, unified autoregressive video models employ 3D rotary position embeddings. MM-RoPE~\cite{yuan2025lumos} scales latent coordinates as $(t,h,w)\times(4,8,8)$, and distributes rotational dimensions across temporal, height, and width axes.  
With these scaled indices, masked-cache attention is computed as
\begin{equation}
\resizebox{0.88\linewidth}{!}{$
\operatorname{Attn}(Q,K,V)=
\operatorname{softmax}\!\left(
\frac{Q\, R_{\Theta,T,H,W}^{d} K^{\top}}{\sqrt{d}}
+ M_{\text{causal}}
\right)V,$}
\label{eq:mmrope_attention_masked_kv}
\end{equation}
where $R_{\Theta,T,H,W}^{d}$ is the 3D rotary operator and $M_{\text{causal}}$ enforces intra-frame bidirectionality and inter-frame temporal causality.


\begin{figure*}[t]
  \centering
  \includegraphics[width=1.0\textwidth]{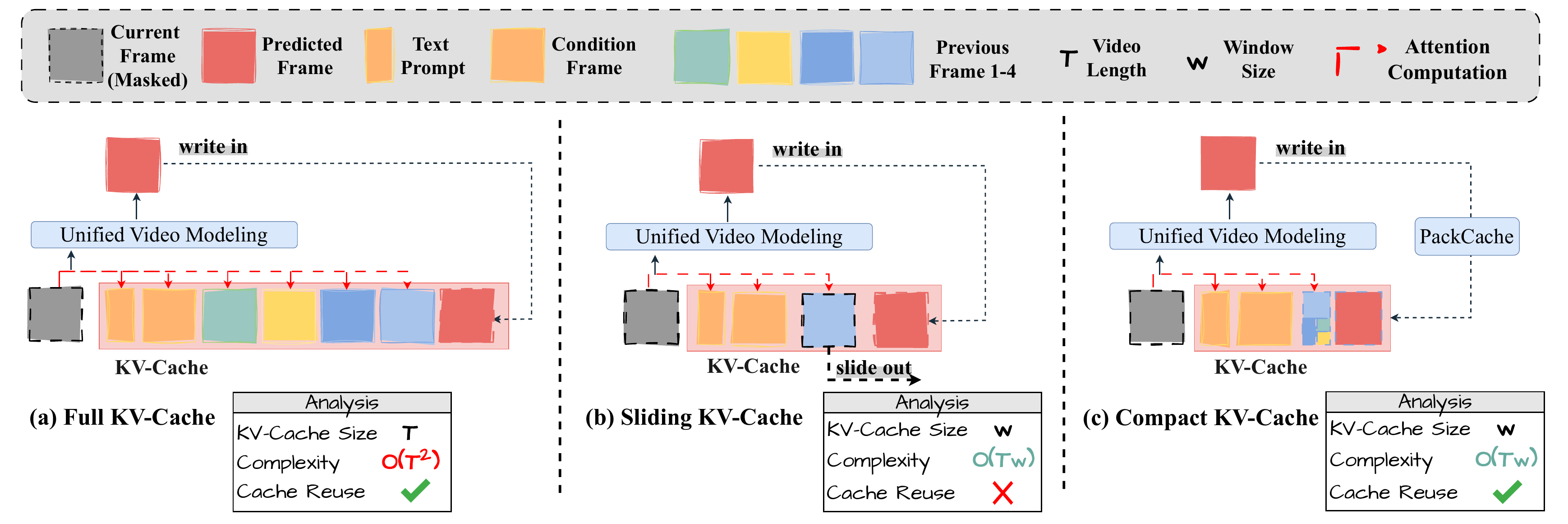}
  \vspace{-0.2in}
  \caption{Comparison of KV-cache strategies in unified autoregressive video generation. At each timestep, the model predicts the current masked frame by attending to the text prompt, conditioning image, and cached previous frames. Once all tokens of the current frame are predicted, the resulting latent frame is written into the KV cache. (a) \textbf{Full KV-Cache} stores all previously generated frames, enabling complete cache reuse but resulting in a KV-cache that grows linearly with video length $T$, leading to quadratic attention cost. (b) \textbf{Sliding KV-Cache} keeps only the most recent $W$ frames, reducing complexity to $\mathcal{O}(TW)$ but sliding out older frames and thus losing long-range temporal context. (c) \textbf{Compact KV-Cache (ours)} compacts previous frames based on their spatiotemporal relevance, retaining only informative tokens while staying within the same window budget $W$. This preserves long-range context through effective cache reuse while maintaining the same $\mathcal{O}(TW)$ complexity as the sliding-window approach.}

  \vspace{-0.2in}
  \label{fig:compare}
\end{figure*}

\subsection{Analyzing Attention Dynamics of KV Cache}
\label{subsec:analysis}
We analyze the attention attribution and temporal dynamics in the video generation process, since understanding how current-frame tokens interact with the cached history is crucial for designing effective KV-cache management strategies.
Attention Heatmaps Fig.~\ref{fig:short-a} and Fig.~\ref{fig:short-d} visualize the attention heatmap of the decoder during the generation of the third latent frame. The horizontal axis is divided into sequential regions corresponding to the \emph{text prompt}, \emph{condition image}, \emph{previous frame~1}, \emph{previous frame~2}, and the \emph{current frame~3} being generated. The vertical axis represents query tokens within the current frame. Based on this partitioning, we analyze how each decoder layer allocates attention across the different information sources described above. Figs.~\ref{fig:short-b}--\ref{fig:short-c} and Figs.~\ref{fig:short-e}--\ref{fig:short-f} presents quantitative attention weights averaged across heads and tokens for both a shallow layer (L0) and a deep layer (L15). 

Three major patterns can be observed:
\begin{itemize}
  \item \textbf{Condition Anchoring.}
  Across layers and diffusion steps, the text prompt and conditioning image consistently retain a non-negligible share of attention.  
  They function as persistent semantic and appearance anchors, remaining attended even when deeper layers shift focus toward temporal context.
  \item \textbf{Cross-frame Attention Decay.}
  Attention to previous frames decreases monotonically with temporal distance, consistently favoring nearer frames over farther ones.  
  This near-stronger-than-far pattern appears across layers and steps, reflecting the model’s inherent temporal locality bias.
  \item \textbf{Step-wise Evolution.}
  Across generation steps, attention naturally shifts from relying on previous context to focusing on current-frame tokens for spatial refinement. Early steps draw more from past frames, while later steps—especially in deeper layers such as L15—are dominated by current-frame self-attention.
\end{itemize}





\subsection{Sliding-Window for AR Video Generation}
\label{subsec:limitation}
To maximize efficiency, the KV-cache budget is constrained to the number of tokens contained in a single latent frame (all operations occur in latent space\cite{nvidia2025cosmosworldfoundationmodel}). A straightforward approach is to adopt a sliding-window mechanism (See Fig.\ref{fig:compare}). In large language models, sliding-window inference keeps only a fixed number of recent tokens in the KV cache, discarding older ones to bound attention cost and mitigate the attention-sink effect \cite{xiao2024efficient}. Similarly, in video generation, the sliding window operates at the frame level, preserving only the tokens of the most recent latent frame- removing earlier ones while still keeping the semantic anchors (the text prompt and conditioning image). However, this strategy suffers from three fundamental drawbacks: 
(1) it discards all earlier $(K,V)$ entries, preventing any temporal reuse; 
(2) the lack of long-range history weakens temporal coherence and leads to drift in long-video generation; and 
(3) even the retained frame is redundant, containing many masked tokens that waste cache capacity without contributing to attention. These limitations highlight the need for a more effective KV-cache management strategy, motivating our approach to retain long-range information while controlling cache size.

\subsection{PackCache} 
\label{subsec:packcache}
\textbf{PackCache.} PackCache addresses the KV-cache bottleneck through selective token retention guided by spatiotemporal relevance. Rather than uniformly discarding previous frames or naively applying sliding windows, our method strategically allocates a fixed cache budget across multiple previous frames according to their temporal distance and semantic importance. 

\begin{figure*}[t]
  \centering
  \includegraphics[width=1.0\textwidth]{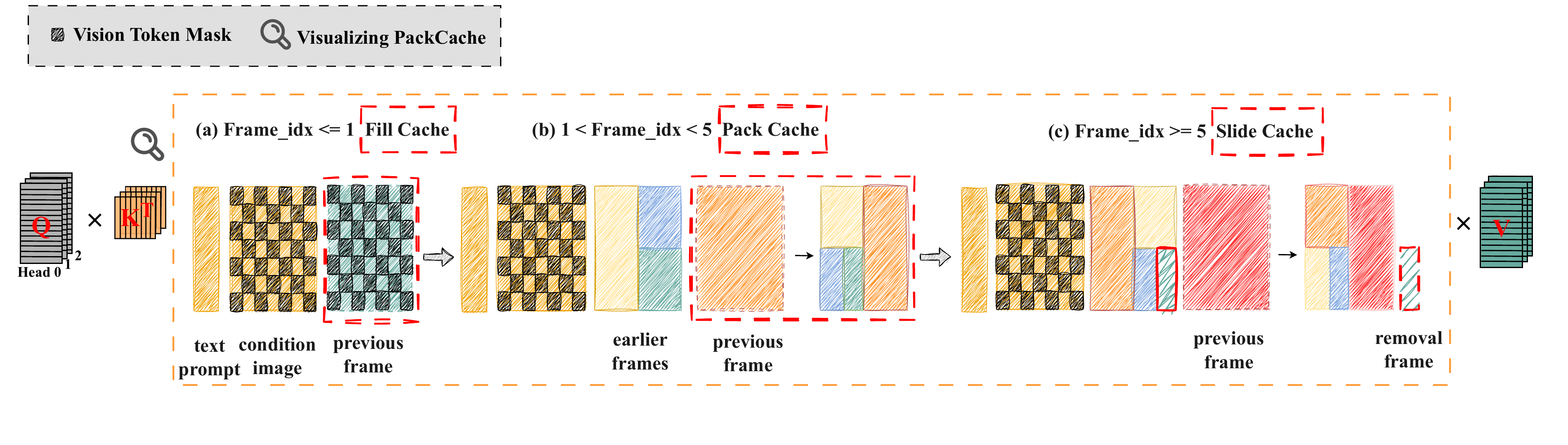}
    \vspace{-0.2in}
  \caption{PackCache workflow. \textbf{(a) Fill Cache}: Incoming frames are appended until the KV cache reaches its capacity. Text-prompt and conditioning-image tokens are stored as persistent semantic anchors. \textbf{(b) Pack Cache}: Once the cache becomes full, the history frames are compacted by removing masked tokens and reallocating per-frame budgets based on the attention-decay kernel $g(d)=\rho^{d}$. (Unlike standard masked KV caching, PackCache directly removes masked positions rather than storing them as zeros.) \textbf{(c) Slide Cache}: As generation proceeds, new frames are inserted and the oldest frames are removed, with each retained frame repacked under the same allocation rule. For visualization, KV tokens are reshaped into 2D grids.}

  \vspace{-0.2in}

  \label{fig:method}
\end{figure*}

Let the latent video tokens of frame $t$ be
\(
X_t = \{x_{t,1}, \dots, x_{t,N}\},
\)
where each token \(x_{t,i} \in \mathbb{R}^{d_x}\) is a $d_x$-dimensional latent feature vector.
The corresponding key/value representations are $K_t\in\mathbb{R}^{N\times d_k}$ and
$V_t\in\mathbb{R}^{N\times d_v}$, where each row of \(K_t\) and \(V_t\) contains the key and value embedding of a token in \(X_t\), respectively.
At each generation step, the KV cache stores a subset of previous frames:
\begin{equation}
\mathcal{C}_t=\{K_{t-d},V_{t-d}\}_{d=1}^{D_t},
\end{equation}
where $D_t \le W$ and $W$ is the window capacity. As illustrated in Fig.~\ref{fig:method}, PackCache operates in three regimes:
(\textit{i}) \textbf{Fill Cache} ($D_t<W$), where new frames are appended
until the budget is full; (\textit{ii}) \textbf{Pack Cache} ($D_t=W$),
where existing cache entries are compacted to remain within the fixed budget;
and (\textit{iii}) \textbf{Slide Cache} ($t>W$), where the cache is maintained
around a moving temporal horizon under the same packing rule.

\noindent\textbf{Fill-Cache Stage with Condition Anchoring.}
In the early stage ($D_t < W$), the cache is populated following the principle of
\emph{condition anchoring}: textual and visual conditioning inputs (the prompt
and conditioning image) act as semantic anchors rather than temporal elements.
Their contributions do not decay with temporal distance, unlike previous
frames whose importance decreases across time.  
Therefore, conditioning entries are excluded from the temporal decay distribution (Eq.~(\ref{eq:norm_decay})) and are each allocated a fixed quota
$\gamma_{\text{text}}$ and $\gamma_{\text{cond}}$ of the total KV-cache
budget.  
This ensures that the conditioning keys/values remain persistently accessible in
$\mathcal{C}_t$, while temporal pruning and packing are applied solely to
previous frame caches $\{K_{t-d}, V_{t-d}\}$. Note that the first generated frame is added to the KV cache without pruning, since no packing condition is triggered yet.

\noindent\textbf{Pack-Cache Stage with Cross-Frame Attention Decay.} 
Based on the observation that cross-frame attention decays with temporal distance, PackCache models the retention of previous tokens directly through their attention decay profiles. 
Let $\mu_d$ denote the mean attention score of the $d$-th previous frame (with $d=1$ being the most recent frame and larger $d$ indicating more distant history). Empirically, $\mu_d$ decreases with temporal distance as shown in Fig.~\ref{fig:heatmap}, suggesting that the relative importance of previous frames can be approximated by a decaying function. 
PackCache therefore models temporal importance using an exponential form
\begin{equation}
\mu_d = C e^{-\alpha d} = C\rho^{d}, \qquad \rho = e^{-\alpha},
\end{equation}
where $C$ is a normalization constant, $\alpha>0$ controls the decay rate, and $\rho\in(0,1)$ is the corresponding decay factor. 
This yields a decay kernel
\begin{equation}
g(d)=\rho^{d}.
\label{eq:decay}
\end{equation}

Given a temporal window of $W$ history frames (i.e., at most $W$ previous frames are kept active), this kernel is converted into a
normalized KV-cache allocation
\begin{equation}
b_d = \frac{g(d)}{\sum_{j=1}^{W} g(j)}, \qquad
\sum_{d=1}^{W} b_d = 1,
\label{eq:norm_decay}
\end{equation}
where $b_d$ denotes the fraction of the total token budget assigned to the $d$-th history frame.
This ensures that each frame $t-d$ contributes no more than $b_d$ of the total
token capacity.
During the Pack-Cache stage (with history depth $D_t = W$), tokens in
$\mathcal{C}_t$ (the KV cache at time $t$) are compacted to meet these per-frame budgets.
Crucially, PackCache removes all masked positions when storing history frames, keeping only unmasked and attention-relevant tokens. 
This attention-guided, geometrically decayed allocation produces a compact KV-cache that preserves long-range temporal structure while eliminating redundant masked entries under a fixed memory budget. 

To prevent distant frames from vanishing completely, we optionally impose a minimum quota $b_{\min}$; if $Wb_{\min}>1$, First In First Out truncation reduces the active window size.

\noindent\textbf{Simplified Closed-Form Decay for Deployment.}
Empirically, we find that a one-frame half-life best matches the behavior of large autoregressive video models (i.e., $\mu_{d+1}\approx \tfrac12\mu_d$). Setting the decay factor to $\rho=\tfrac12$ in Eq.~(\ref{eq:decay}), yielding a simple and stable closed-form allocation:
\begin{equation}
b_d = 2^{-\min(d,\,W-1)}, \qquad d=1,\dots,W.
\label{eq:final_form}
\end{equation}
This produces the intuitive pattern
\[
[1],~~\left[\tfrac12,\tfrac12\right],~~
\left[\tfrac12,\tfrac14,\tfrac14\right],~~
\left[\tfrac12,\tfrac14,\tfrac18,\tfrac18\right],\ldots
\]
and analytically satisfies $\sum_{d=1}^{W}b_d=1$.  
The packed token budget becomes
\begin{equation}
t_d = B_{\text{one}} \, b_d,
\qquad \sum_{d=1}^{W} t_d = B_{\text{one}},
\end{equation}
where $t_d$ is the number of tokens assigned to the $d$-th history frame and $B_{\text{one}}$ is the full-token count of one latent frame.
This temporally aware packing maintains the KV-cache at a constant cost, reuses multiple previous frames with higher fidelity for nearby ones, and significantly reduces temporal drift while preserving long-range semantic consistency.

\noindent\textbf{Spatially Preserving Position Embedding.} Unified autoregressive video generation models use mixed 1D--3D rotary position embeddings
(MM-RoPE\cite{yuan2025lumos}), where global sequence indices follow 1D RoPE and
visual tokens adopt factorized 3D coordinates $(t,h,w)$.
Under a sliding-window KV-cache, temporal indices become discontinuous because
only recent frames are retained, which disrupts MM-RoPE's relative positional
structure. To maintain positional consistency, we apply a lightweight \emph{rebase}
operation whenever the window advances.  
The 1D global index is shifted to remain continuous, while in the 3D index only
the temporal component is updated:
\begin{equation}
\resizebox{0.85\linewidth}{!}{$
(pos_t,\, pos_h,\, pos_w) \leftarrow (pos_t - \Delta_t,\, pos_h,\, pos_w)
$}
\end{equation}

\noindent where $\Delta_t$ corresponds to the number of dropped frames. By keeping $(pos_h,pos_w)$ fixed, the spatial layout is preserved, whereas the
rebased $pos_t$ restores temporal continuity within the window. 

\noindent\textbf{High-level Summary of PackCache:} The core mechanism of PackCache operates in three stages (Fig.\ref{fig:method}): during the \textbf{Fill-Cache} stage, we populate the cache while dedicated quotas for text prompts and conditioning images fixed—these semantic anchors receive persistently high attention and are therefore exempted from temporal decay. 
Once the cache reaches capacity, the Pack-Cache stage compacts previous frames by modeling attention decay as an exponential function of temporal distance, allocating exponentially fewer tokens to older frames while explicitly removing all masked positions. 
For extended generation, the \textbf{Slide-Cache} stage maintains this temporally-weighted packing strategy within a moving window. Critically, to preserve the spatial structure required by 3D rotary position embeddings (MM-RoPE), we introduce a \textbf{spatially preserving rebasing} operation that maintains continuous temporal indices and coherent $(h, w)$ coordinates despite token eviction. This training-free approach enables stable long-video generation under strict memory constraints by balancing temporal coverage with per-frame fidelity.

\begin{figure*}[t]
  \centering
  \includegraphics[width=1.0\textwidth]{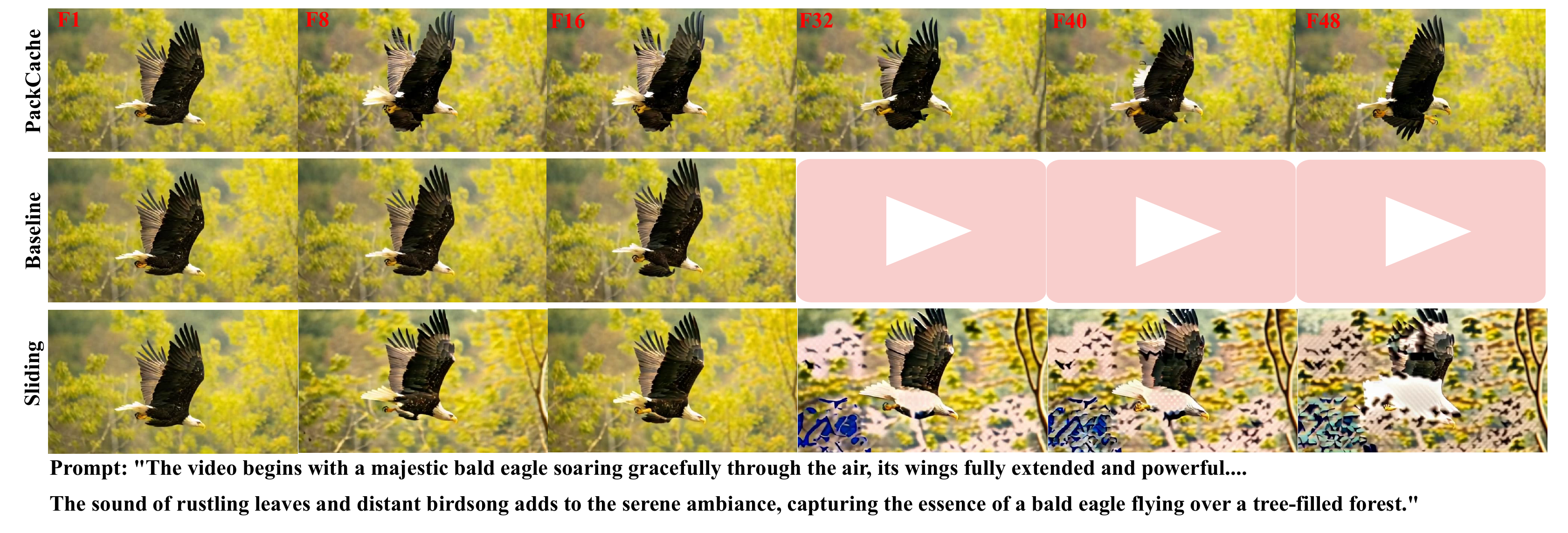}
  \vspace{-0.2in}
  \caption{48-frame video generation results using the 3B Lumos-1 on A40 GPU. \textbf{PackCache} produces temporally stable and coherent motion across the entire sequence, whereas the \textbf{Sliding Window} exhibits severe drift and structural degradation in the later frames, demonstrating PackCache's advantage in long-horizon video generation.}
  \label{fig:vis}
\end{figure*}


\section{Experiment}
\label{sec:experiment}

\subsection{Setup}
\noindent\textbf{Models.}
We conduct all experiments using the 3B Lumos-1 model~\cite{yuan2025lumos} at a target resolution of $672\times384$. Lumos-1 is, to the best of our knowledge, the only open-source unified autoregressive video generator. After tokenization with the Vision Tokenizer~\cite{nvidia2025cosmosworldfoundationmodel}, each frame is compressed into 4,084 visual tokens, with the text prompt contributing a few hundred additional tokens. To study efficiency across temporal scales, we benchmark two configurations: a short-video setting with 7 latent frames (24 video frames) and a long-video setting with 13 latent frames (48 video frames).


\noindent\textbf{Hardware.} Experiments are conducted on NVIDIA A40 (48,GB) and H200 (141,GB) GPUs. We report end-to-end latency and speed-up on both platforms to evaluate the efficiency and scalability across GPU generations.

\noindent\textbf{Dataset.} For evaluation, we construct a 160-image subset sampled from the I2V portion of VBench \cite{huang2024vbench}, selecting images uniformly at random. Following the default configuration of Lumos-1~\cite{yuan2025lumos}, each image is paired with a rewritten caption generated by the Qwen-32B model \cite{yang2025qwen3}, which serves as the text prompt for image-to-video generation.  

\noindent\textbf{Baselines.} We compare our approach against two baselines: (1) the original Lumos-1 model with the \emph{full} KV-cache, and (2) a \emph{sliding-window} variant that retains only the most recent latent frame in the cache. Both the sliding-window and our PackCache operate under a nearly identical KV-cache budget of approximately 8{,}417 tokens (with minor variation due to prompt length). 
Unlike sliding-window methods that discard all older memory, PackCache adds only a small number of meta tokens—on the order of a few tens—to encode frame-boundary information and enforce minimum-quota packing. This negligible overhead enables far richer temporal retention than simple truncation.

\begin{table*}[t]
\centering
\small
\begin{tabular}{c|l|ccccccc|c}
\toprule
\textbf{Frames} & \textbf{Method} & \textbf{Subj. $\uparrow$} & \textbf{Backg. $\uparrow$} & \textbf{Motion. $\uparrow$} & \textbf{Aesth. $\uparrow$} & \textbf{I2V-Sub. $\uparrow$} & \textbf{I2V-Back. $\uparrow$} & \textbf{DynDeg. $\uparrow$} & \textbf{Overall} \\
\midrule
\multirow{3}{*}{24} 
 & Baseline       & 0.973 & 0.960 & 0.988 & 0.600 & 0.969 & 0.974 & 0.019 & 0.783 \\
 & Sliding Window  & 0.931 & 0.948 & 0.982 & 0.570 & 0.948 & 0.954 & 0.081 & 0.773 \\
 & PackCache (ours)      & 0.948 & 0.956 & 0.986 & 0.556 & 0.940 & 0.948 & 0.431 & 0.824 \\
\midrule
\multirow{3}{*}{48} 
 & Baseline       
 & \multicolumn{8}{c}{\cellcolor{red!10}\textbf{OOM}} \\
 & Sliding Window  & 0.860 & 0.906 & 0.979 & 0.549 & 0.924 & 0.936 & 0.031 & 0.741 \\
 & PackCache (ours)      & 0.891 & 0.920 & 0.983 & 0.550 & 0.915 & 0.929 & 0.263 & 0.779 \\
\bottomrule
\end{tabular}
\caption{Quantitive benchmarking results of PackCache and baseline methods on the I2VBench dataset. Metrics include Subject Consistency (Subj), Background Consistency (Backg), Motion Smoothness (Motion), Aesthetic Quality (Aesth), I2V-Subject (I2V-Sub), I2V-Background (I2V-Back), and Dynamic Degree (DynDeg). For short videos (24 frames), PackCache achieves comparable overall quality to the baseline while additionally improving motion dynamics. For longer videos (48 frames), baseline is out of memory, compared to the sliding-window that retains only the most recent frame tokens, PackCache delivers consistently superior quality, demonstrating better temporal stability and scalability.}
\label{tab:quality}
\end{table*}

\subsection{Quality Evaluation}
Table~\ref{tab:quality} summarizes the quantitive results on the I2VBench subset. For short videos (24 frames), PackCache delivers results on par with the full-cache baseline while offering significantly faster inference. The higher Dynamic Degree metric is consistent with PackCache’s more localized and sparser attention usage. 
For long videos (48 frames), the full-cache baseline cannot complete inference due to OOM even on an H200 GPU. Under the same KV-cache budget, PackCache consistently surpasses the sliding-window baseline across nearly all metrics, as it preserves a richer and more informative subset of historical context, whereas the sliding-window approach retains only the most recent frame. Qualitative comparisons further show smoother and more stable dynamics from PackCache in motions modeling—such as sustained eagle wing flapping—demonstrating its ability to preserve long-term cues under strict memory constraints.

\subsection{Efficiency Evaluation}
Table~\ref{tab:efficiency} summarizes the latency and speedup achieved by different KV-cache strategies on A40 and H200 GPUs for both 24-frame and 48-frame video generation. Since the full-cache baseline fails to complete 48-frame synthesis on both A40 and H200 (often resulting in OOM), its latency is estimated from partial measurements and reported as a conservative lower bound.

\begin{figure}[h]
  \centering
  \begin{subfigure}{\linewidth}
    \includegraphics[width=1.01\linewidth]{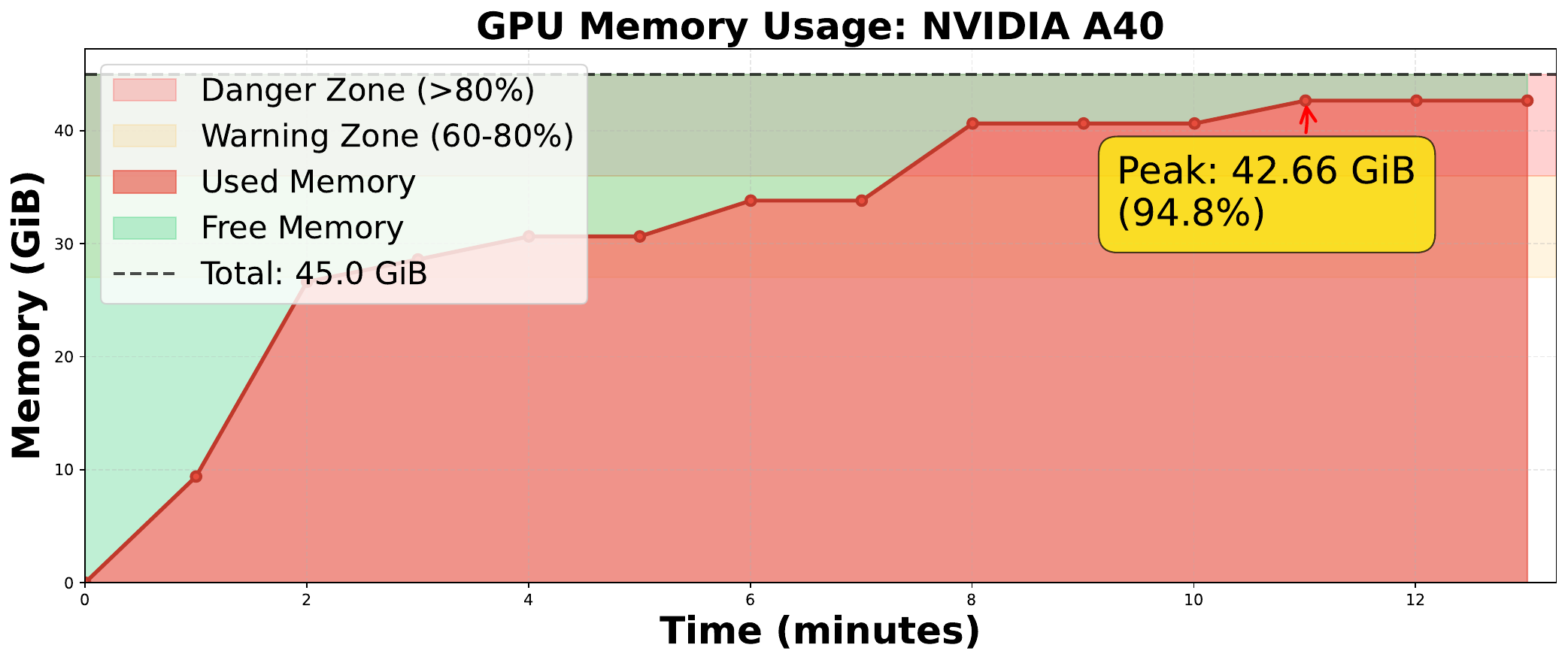}
    \caption{}
    \label{fig:time_ours}
  \end{subfigure}
  \hfill
  \begin{subfigure}{\linewidth}
    \includegraphics[width=1.03\linewidth]{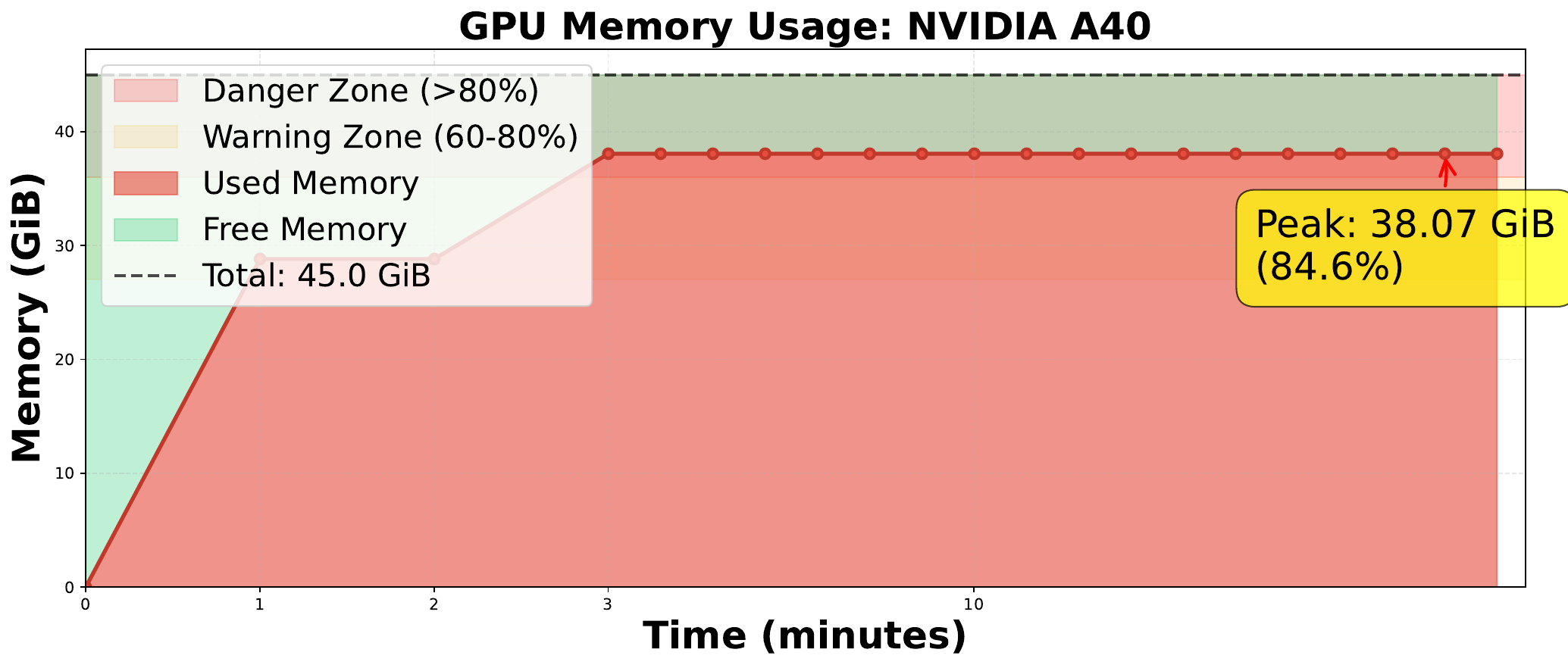}
    \caption{}
    \label{fig:gpu_ours}
  \end{subfigure}
  \vspace{-0.2in}
  \caption{GPU memory usage.\textbf{(a)Baseline} Memory consumption rapidly climbs and peaks at $42.66$\,GiB ($94.8\%$ of a 45\,GiB A40), demonstrating that long-sequence video generation is constrained primarily by KV-cache size. \textbf{(b)PackCache} Video model achieves stable per-frame latency without the linear growth seen in standard autoregressive decoding. GPU memory usage remains bounded (\(\sim\)38\,GiB) throughout generation, independent of video length.}

  \label{fig:overhead_ours}
\end{figure}

Across both GPUs, bounding the KV-cache size yields substantial acceleration. On 24-frame videos, PackCache achieves $1.26\times$ speedup on A40 and $1.45\times$ on H200, approaching sliding-window performance while retaining significantly richer previous context. Sliding-window is only marginally faster due to discarding all old tokens, whereas PackCache introduces a small packing and RoPE-update overhead that is intentionally kept lightweight.
For long videos (48 frames), PackCache delivers $1.75\times$ acceleration on A40 and $2.18\times$ on H200—slightly lower than sliding-window but markedly more stable. Unlike sliding-window, which frequently exhibits temporal drift and structural degradation, PackCache preserves long-range context and remains robust even under strict memory budgets where full-cache decoding is infeasible.
Fig.~\ref{fig:overhead_ours} further illustrates these advantages.

\begin{table}[H]
\centering
\footnotesize
\scalebox{0.9}{
\setlength{\tabcolsep}{1.2pt}
\renewcommand{\arraystretch}{1.3}

\begin{tabular}{c|c|l|cc|cc}
\toprule
\textbf{GPU} & \textbf{Frames} & \textbf{Method} &
\textbf{TOTAL (s)} & \textbf{Speedup↑} &
\textbf{LAST (s)} & \textbf{Speedup↑} \\
\midrule
\multirow{4}{*}{A40}
 & \multirow{2}{*}{24}
   & Baseline        & 731.59 & -       & 163.97 & -        \\
 & & \cellcolor{blue!8}PackCache (ours) 
       & \cellcolor{blue!8}580.84 
       & \cellcolor{blue!8}1.26× 
       & \cellcolor{blue!8}100.53 
       & \cellcolor{blue!8}1.63× \\
\cmidrule(lr){2-7}
 & \multirow{2}{*}{48}
   & Baseline        
       & \textit{2075.45$^{\ast}$} & -       
       & \textit{267.08$^{\ast}$}  & -        \\
 & & \cellcolor{blue!8}PackCache (ours) 
       & \cellcolor{blue!8}1183.99 
       & \cellcolor{blue!8}1.75× 
       & \cellcolor{blue!8}101.15 
       & \cellcolor{blue!8}2.64× \\
\midrule
\multirow{4}{*}{H200}
 & \multirow{2}{*}{24}
   & Baseline        & 201.44 & -        & 46.83 & -         \\
 & & \cellcolor{blue!8}PackCache (ours) 
       & \cellcolor{blue!8}139.40 
       & \cellcolor{blue!8}1.45× 
       & \cellcolor{blue!8}26.89 
       & \cellcolor{blue!8}1.74× \\
\cmidrule(lr){2-7}
 & \multirow{2}{*}{48}
   & Baseline        & 595.33 & -        & 79.02 & -         \\
 & & \cellcolor{blue!8}PackCache (ours) 
       & \cellcolor{blue!8}272.75 
       & \cellcolor{blue!8}2.18× 
       & \cellcolor{blue!8}21.58 
       & \cellcolor{blue!8}3.66× \\
\bottomrule
\end{tabular}}

\caption{Speedup comparison on A40 and H200 for 24-/48-frame generation. 
\textbf{TOTAL} is the full video generation time; \textbf{LAST} is the latency of the final four frames. 
Baseline fails for 48-frame synthesis (OOM), so values are curve-fitted. Overall, our method achieves a 1.6–1.7× speed-up on 24-frame videos and up to 2.6–3.7× acceleration on 48-frame videos across A40 and H200 GPUs.}
\label{tab:efficiency}
\end{table}

\subsection{Ablation Study}
We analyze how the window size $W$ and the minimum quota $b_{\min}$ affect the trade-off between \emph{temporal range} and \emph{per-frame fidelity}.  
A larger window retains more distant history, while a higher $b_{\min}$ enforces a stronger lower bound on the number of tokens preserved for each frame, favoring short-term consistency at the cost of reduced long-range context. Table~\ref{tab:quota} summarizes the performance under different minimum quotas.  We observe a clear trend: too small a quota (e.g., 2-frame equivalent) weakens recent-frame fidelity, leading to degraded reconstruction quality, whereas an overly large quota (4–5 frames) over-allocates to nearby frames and compresses older context prematurely.  The best performance occurs at a 3-frame quota, which provides a balanced allocation across the retained history and yields the highest VBench-i2v score.  

\begin{table}[H]
\centering
\scalebox{0.7}{
\begin{tabular}{l|cccc}
\toprule
\textbf{Minimum Quota} $b_{\min}$ (ratio)          
& $2/W$ & $3/W$ & $4/W$ & $5/W$ \\
\midrule
\textbf{Frame-Equivalent Quota}                 
& 2 Frames & 3 Frames & 4 Frames & 5 Frames \\
\textbf{VBench-i2v} $\uparrow$  & 0.754 & \textbf{0.8326} & 0.8070 & 0.8135 \\
\bottomrule
\end{tabular}}
\caption{Ablation on the minimum quota $b_{\min}$ in PackCache.  
The best performance occurs at a frame-equivalent quota of \textbf{3 frames} 
(i.e., $b_{\min}=3/W$), which balances long-range context and per-frame fidelity. VBench-i2v scores are computed on a small subset.}
\label{tab:quota}
\end{table}

\section{Conclusion}
\label{sec:conclusion}

We presented PackCache, a training-free KV-cache management strategy that substantially improves the efficiency and scalability of unified autoregressive video generation. By analyzing the spatiotemporal structure of visual tokens, we identified three key properties—temporal attention decay, sparse yet stable spatial structure, and persistent conditioning anchors—and translated them into a principled packed-cache design. PackCache compacts KV entries according to their relevance, eliminates masked-token redundancy, and, together with our spatial-preserving positional embedding mechanism, preserves both long-range temporal coherence and intra-frame structure under a fixed memory budget. Experiments on A40 and H200 GPUs show consistent acceleration across both short and long sequences, achieving up to $3.7\times$ speed-up in the most computationally expensive tail frames without sacrificing visual fidelity. 

\clearpage
{
    \small
    \bibliographystyle{ieeenat_fullname}
    \bibliography{main}
}

\clearpage
\setcounter{page}{1}
\appendix
\section*{Appendix}
\label{sec:Supplementary}




\section{Spatially Preserving RoPE}
\begin{figure}[H]
  \centering
  \includegraphics[width=\linewidth]{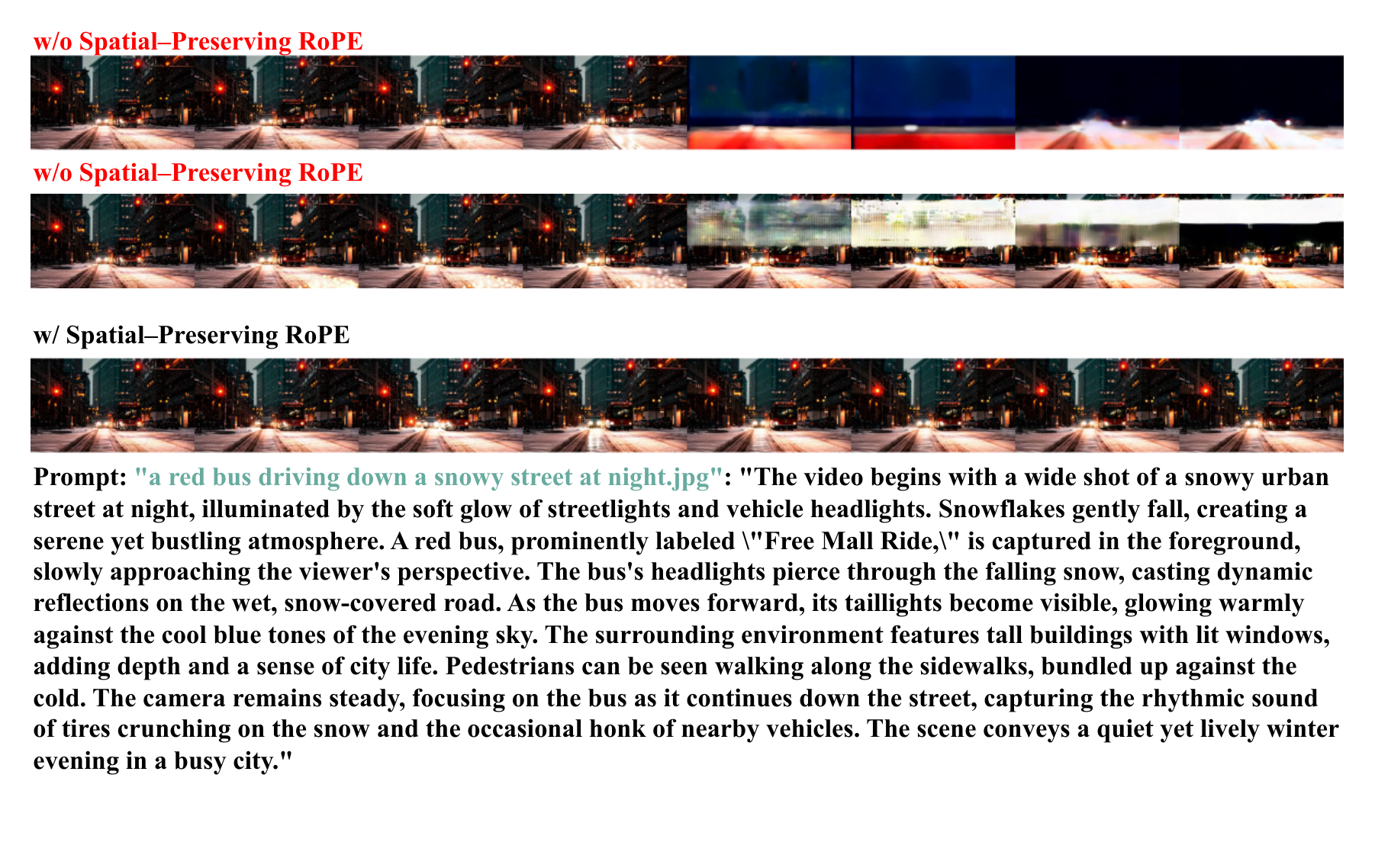}
  \caption{Comparison of different RoPE Strategies. Methods that disrupt spatial indexing (above two rows) introduce frame jitter and structural degradation, especially in later frames. In contrast, Spatial Preserving RoPE maintains spatial integrity by using non-continuous spatial indices and continuous temporal indices, yielding more stable long-sequence generation.}
  \label{fig:rope}
\end{figure}

The Fig.~\ref{fig:rope} compares video generation results with and without spatially preserving RoPE (ours). Our method maintains the structural integrity of the spatial dimensions by using non-continuous indices when computing 3D positional encodings, while retaining continuous indexing along the temporal dimension, which prevents the positional indices exceeding the valid RoPE encoding range as video length increases.
As shown in Fig.~\ref{fig:rope}, the first two rows w/O spatially preserving RoPE —both of which disrupt the spatial indexing structure (with the first row using fully continuous indexing)—exhibit noticeable frame jitter and content degradation, particularly in the later portions of the video. In contrast, with spatially preserving  substantially improves spatial stability and preserves visual coherence across long sequences.

\section{Full-frame Visualizations}
\begin{figure*}[t]
  \centering
  \includegraphics[width=1.0\textwidth]{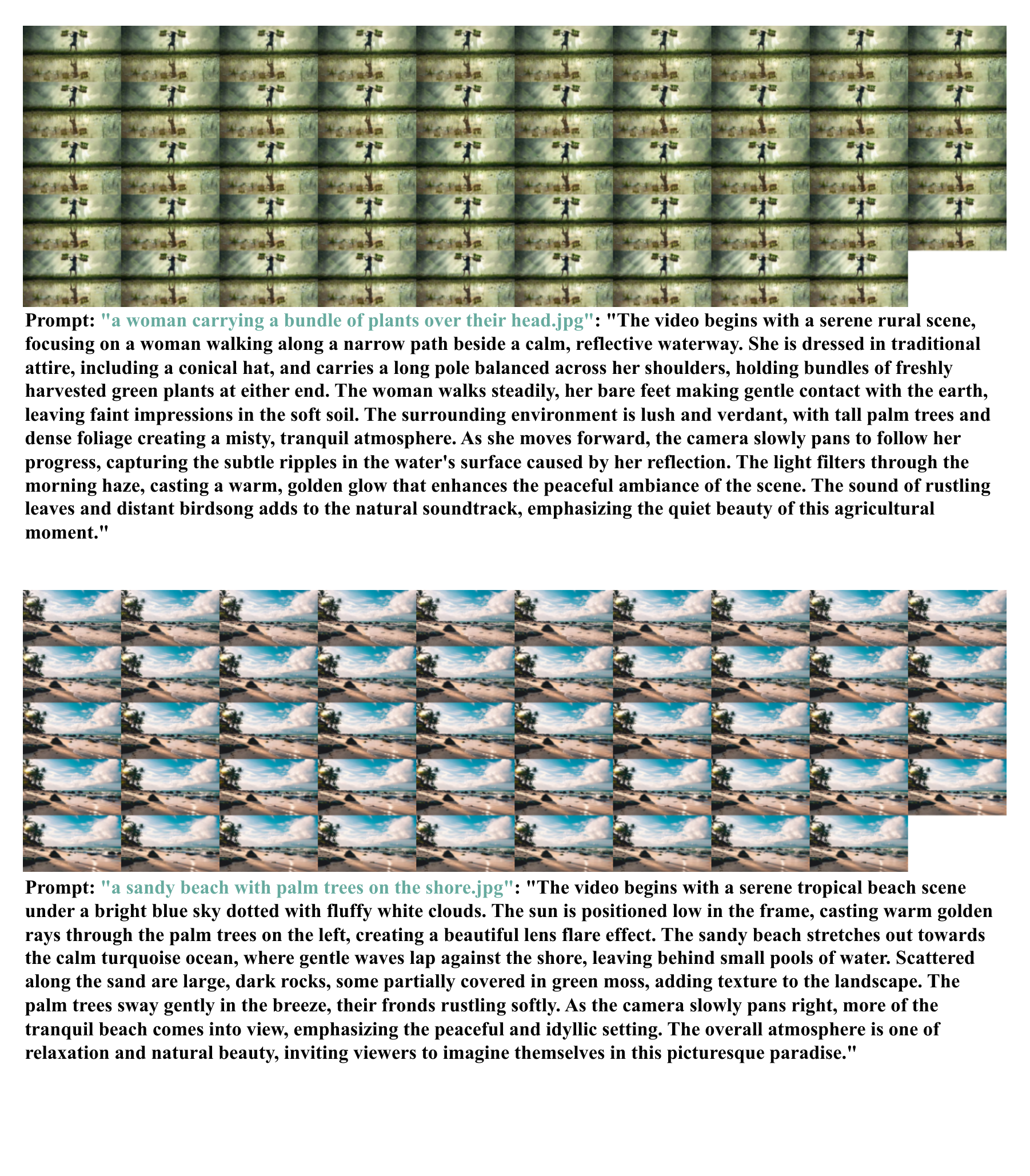}
  \vspace{-0.8in}
  \caption{Full-frame Visualizations.}
  \label{fig:vis_suppl}
\end{figure*}

Fig.~\ref{fig:vis_suppl} present full-frame visualizations of all generated video frames (48 frames per video) without any cropping or post-processing. These results span a diverse set of categories, including humans, landscapes, plants, animals, architecture, and fluid dynamics.

\vspace{0.6in}

\section{Additional Attention Heatmaps}
\begin{figure}[H]
  \centering
  
  \begin{subfigure}{\linewidth}
    \centering
    \includegraphics[width=\linewidth]{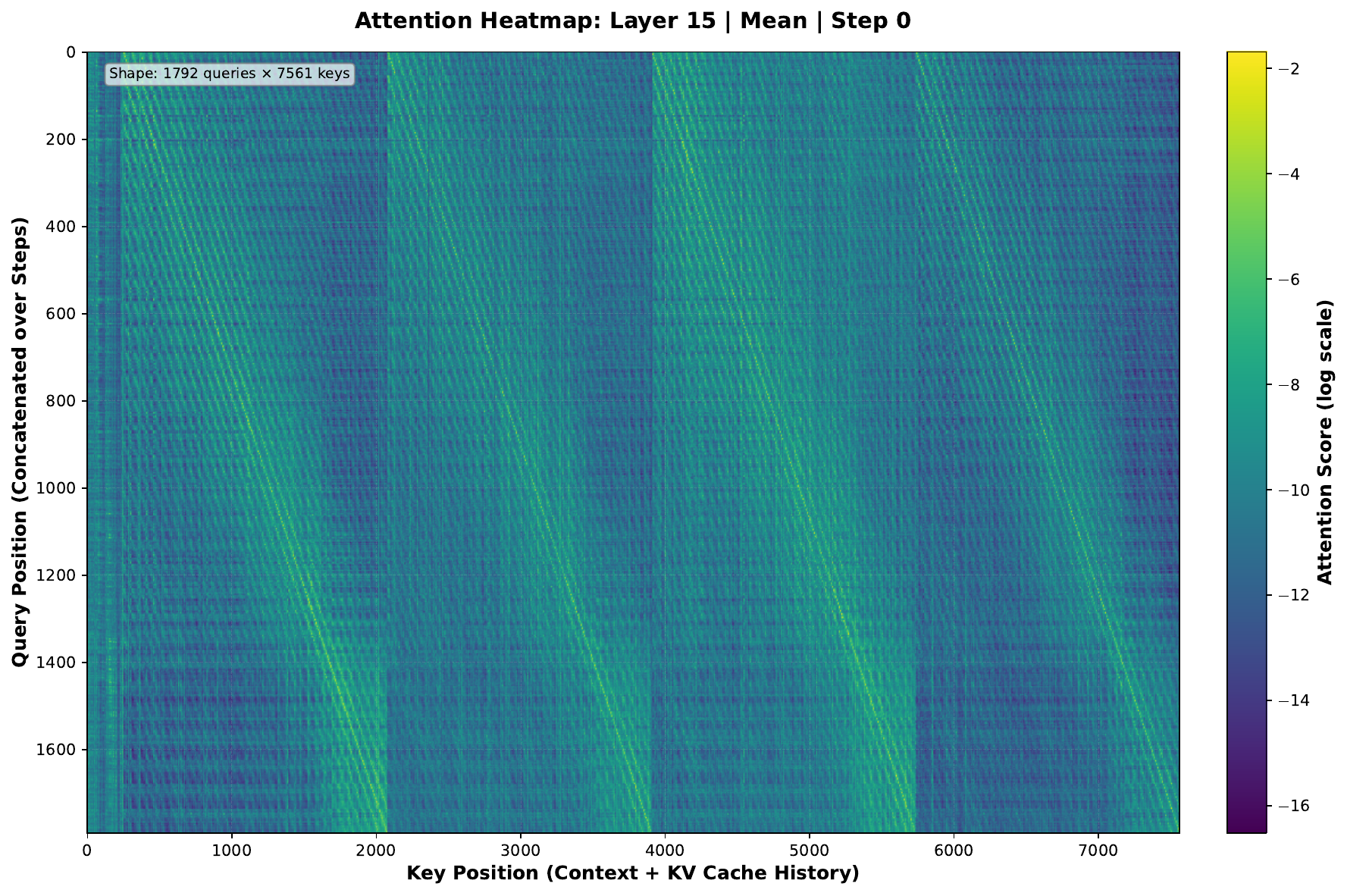}
  \end{subfigure}
  \begin{subfigure}{\linewidth}
    \centering
    \includegraphics[width=\linewidth]{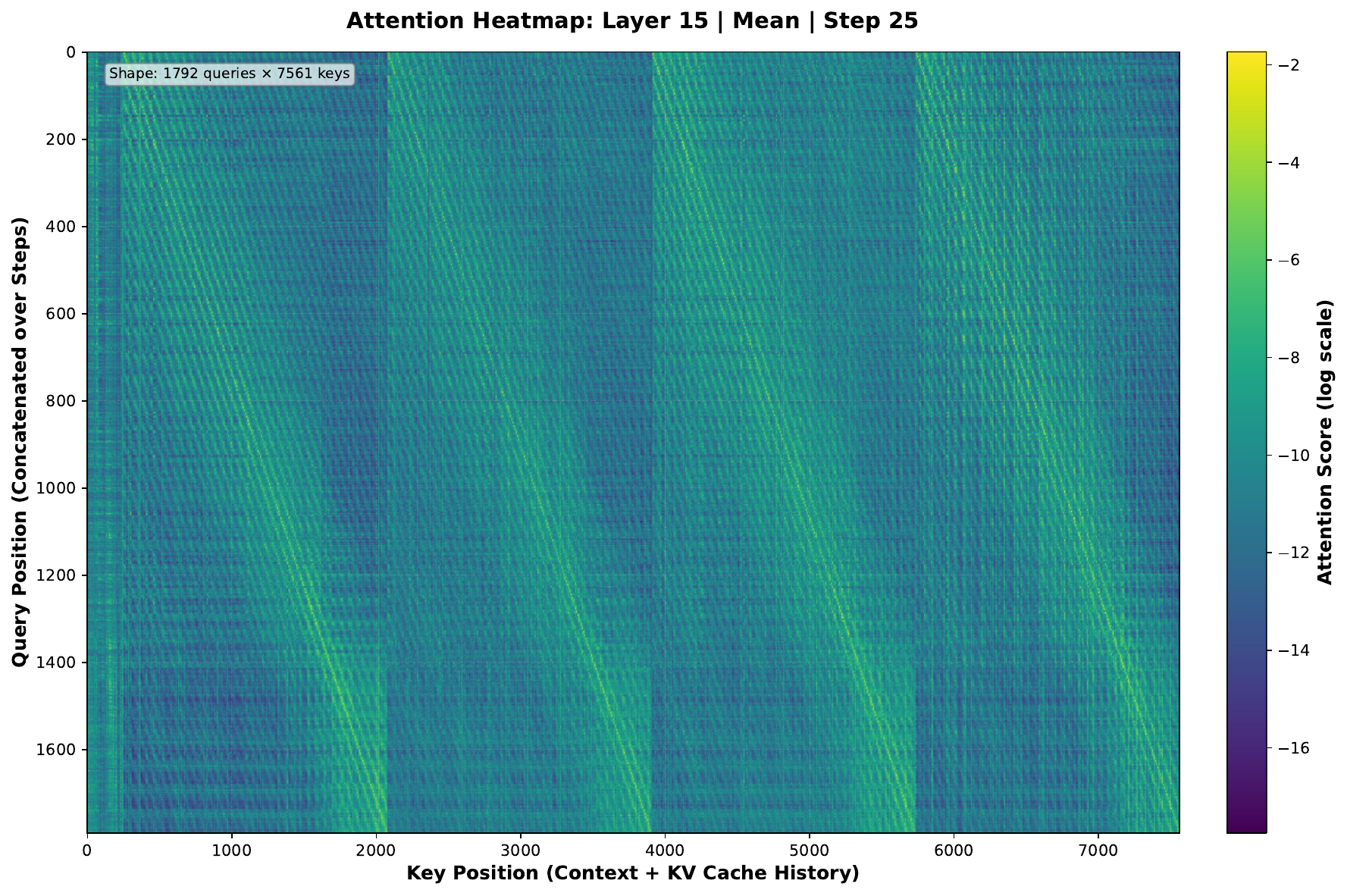}
  \end{subfigure}
  \begin{subfigure}{\linewidth}
    \centering
    \includegraphics[width=\linewidth]{graph/heatmap/frame_3_step_49_layer_15.pdf}
  \end{subfigure}

  \caption{
  Additional attention heatmaps for Layer~15 of Lumos-1. We show the evolution of cross-frame attention patterns at three autoregressive timesteps: Step~0, Step~25, and Step~49.}
\end{figure}

\begin{figure*}[t]
  \centering
  \includegraphics[width=1.0\textwidth]{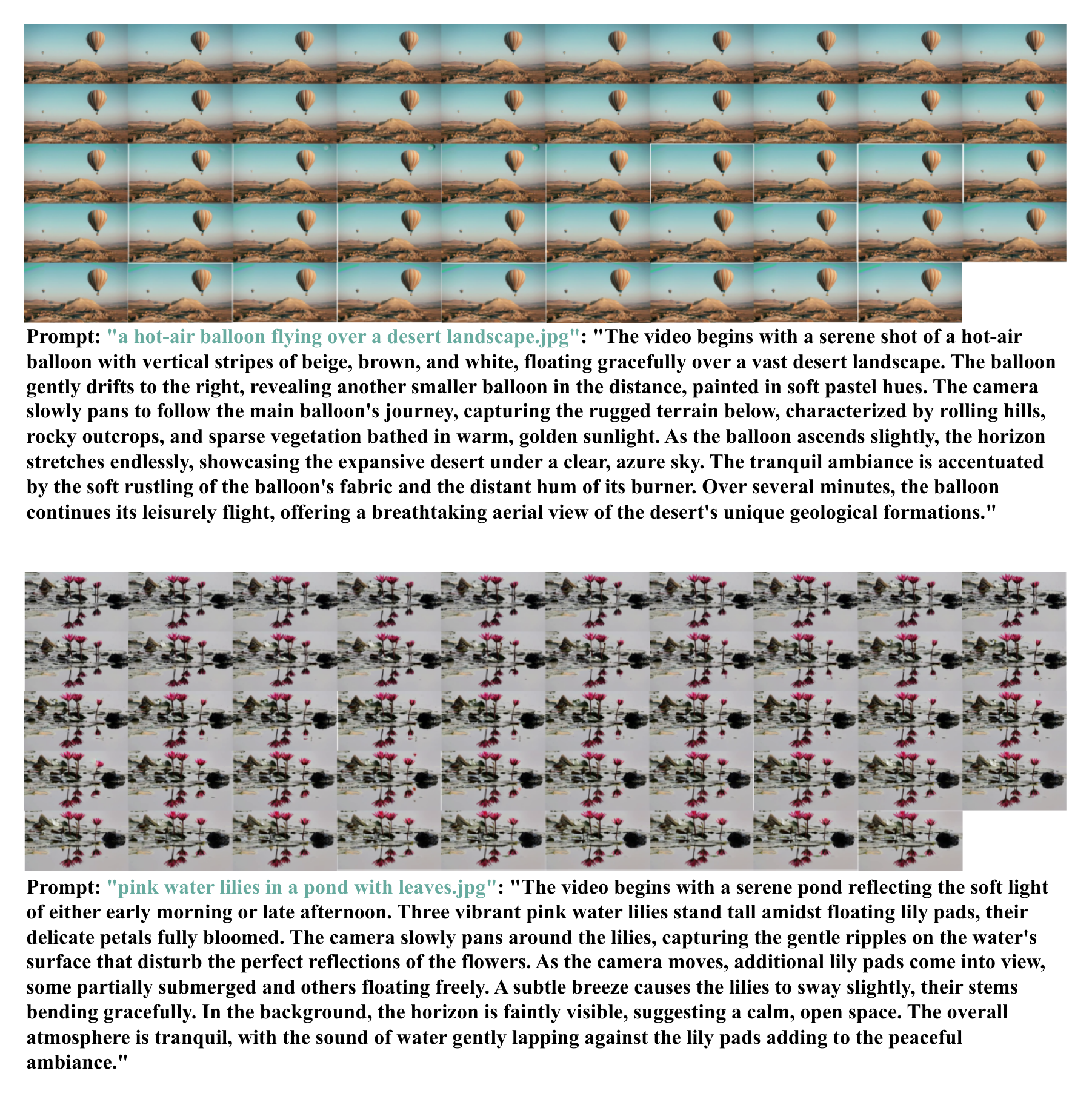}
\end{figure*}

\begin{figure*}[t]
  \centering
  \includegraphics[width=1.0\textwidth]{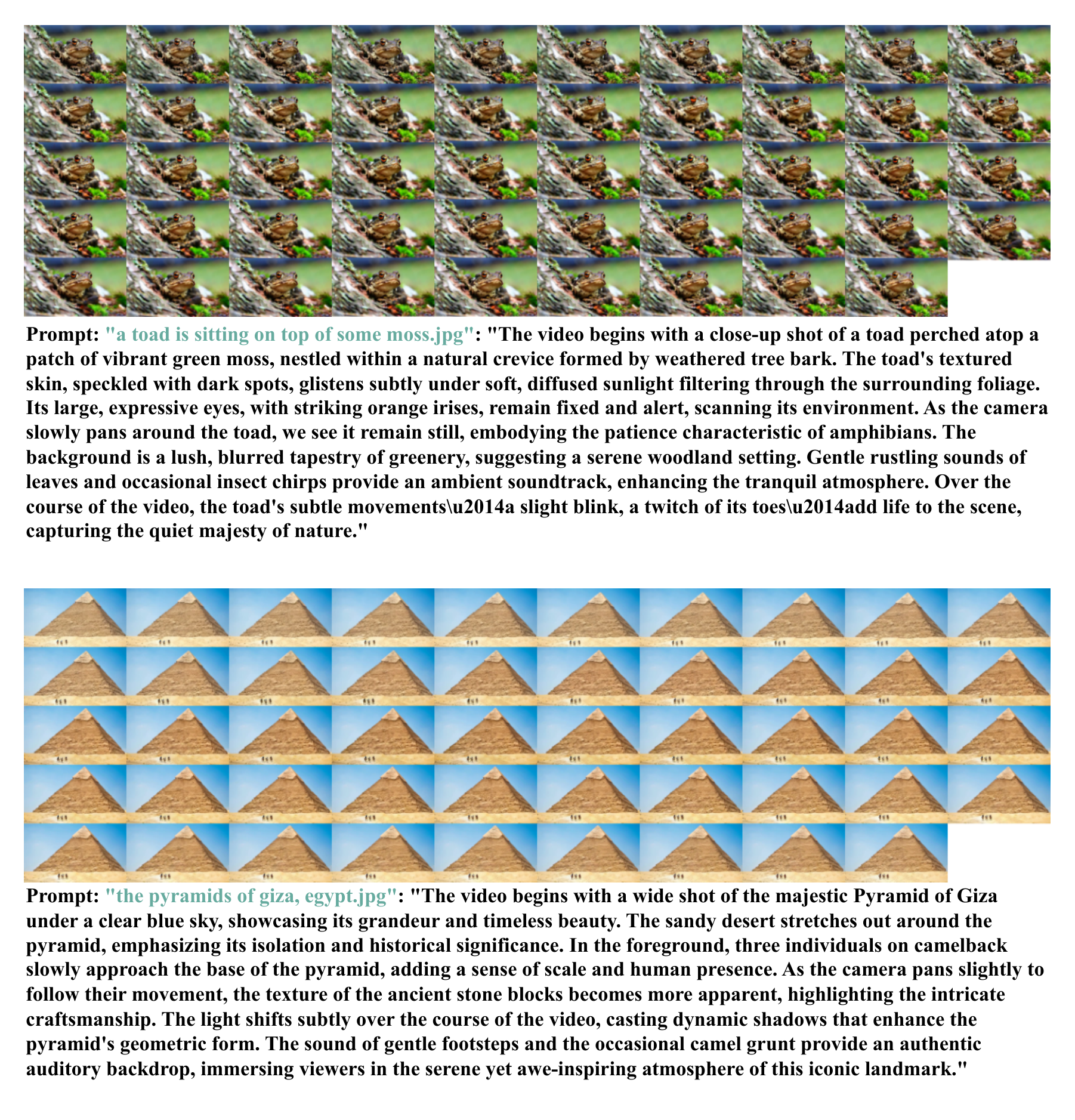}
\end{figure*}

\begin{figure*}[t]
  \centering
  \includegraphics[width=1.0\textwidth]{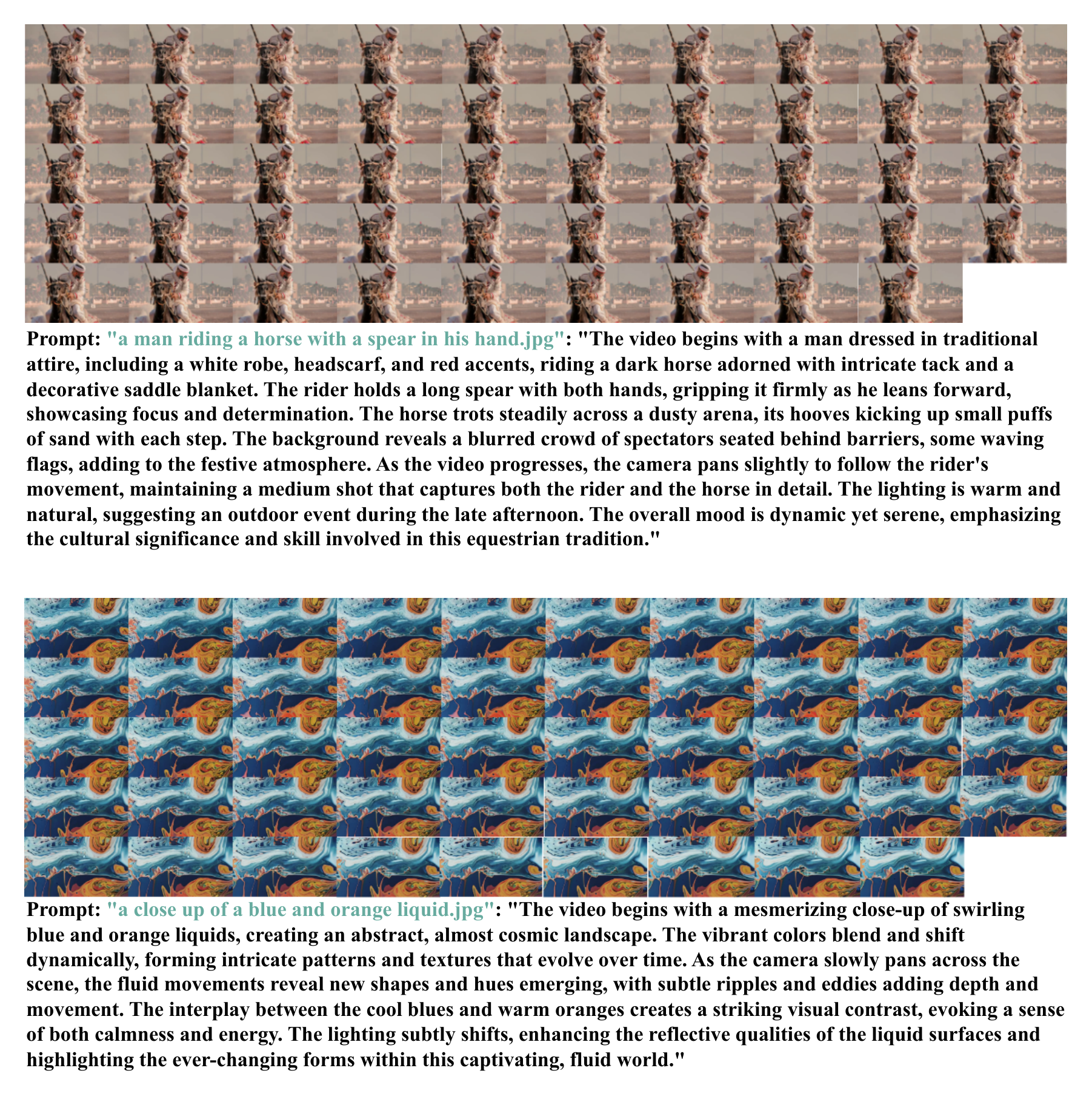}
\end{figure*}

\end{document}